%% file: cas-sc-template.tex
\documentclass[a4paper,fleqn]{cas-sc}

\usepackage{cite}
\usepackage{amsmath,amsfonts,amssymb,booktabs}
\usepackage{algorithm}
\usepackage{algorithmic}
\usepackage{array,multirow}
\usepackage{fixltx2e}
\usepackage{stfloats}
\usepackage{url}
\usepackage{setspace}
\usepackage{enumerate}
\usepackage{amssymb}
\usepackage{amsmath}
\usepackage[switch]{lineno}     
\usepackage{booktabs}           
\usepackage{multirow}           
\usepackage{array}              
\usepackage{threeparttable}     
\usepackage{color}              
\usepackage{float}              
\usepackage{makecell}           
\usepackage{algorithm}          
\usepackage{subfigure}
\usepackage[numbers,sort&compress]{natbib}
\usepackage[justification=centering]{caption}		
\usepackage{bm}
\usepackage{bbding}
\usepackage{makecell}
\usepackage{soul}
\soulregister{\cite}7 
\soulregister{\citep}7 
\soulregister{\citet}7 
\soulregister{\ref}7 
\soulregister{\pageref}7 

\def\tsc#1{\csdef{#1}{\textsc{\lowercase{#1}}\xspace}}
\tsc{WGM}
\tsc{QE}

\begin{document}
\let\WriteBookmarks\relax
\def\floatpagepagefraction{1}
\def\textpagefraction{.001}

\shorttitle{}    

\shortauthors{}  

\title [mode = title]{Faster OreFSDet : A Lightweight and Effective Few-shot Object Detector for Ore Images}  

\tnotemark[1] 

\tnotetext[1]{Corresponding author.}

%

\author[1,2,3]{Yang~Zhang}[]
\author[1,2]{Le~Cheng}[]
\author[1,2]{Yuting~Peng}[]
\author[4]{Chengming~Xu}[]
\author[4]{Yanwei~Fu}[]
\author[5]{Bo~Wu}[]
\author[1,2]{Guodong~Sun}[]
\cormark[1]






\affiliation[1]{organization={School of Mechanical Engineering},
            addressline={Hubei University of Technology}, 
            city={Wuhan},
            postcode={430068}, 
            country={China}}
\affiliation[2]{organization={Hubei Key Laboratory of Modern Manufacturing Quality Engineering},
            addressline={Hubei University of Technology}, 
            city={Wuhan},
            postcode={430068}, 
            country={China}}
\affiliation[3]{organization={National Key Laboratory for Novel Software Technology},
			addressline={Nanjing University}, 
			city={Nanjing},
			postcode={210023}, 
			country={China}}
\affiliation[4]{organization={School of Data Science},
			addressline={Fudan University}, 
			city={Shanghai},
			postcode={200433}, 
			country={China}}
\affiliation[5]{organization={Shanghai Advanced Research Institute},
			addressline={Chinese Academy of Sciences}, 
			city={Shanghai},
			postcode={201210}, 
			country={China}}



\begin{abstract}
For the ore particle size detection, obtaining a sizable amount of high-quality ore labeled data is time-consuming and expensive. 
General object detection methods often suffer from severe over-fitting with scarce labeled data. 
Despite their ability to eliminate over-fitting, existing few-shot object detectors encounter drawbacks such as slow detection speed and high memory requirements, making them difficult to implement in a real-world deployment scenario.
To this end, we propose a lightweight and effective few-shot detector to achieve competitive performance with general object detection with only a few samples for ore images.
First, the proposed support feature mining block characterizes the importance of location information in support features. 
Next, the relationship guidance block makes full use of support features to guide the generation of accurate candidate proposals. 
Finally, the dual-scale semantic aggregation module retrieves detailed features at different resolutions to contribute with the prediction process.
Experimental results show that our method consistently exceeds the few-shot detectors with an excellent performance gap on all metrics. Moreover, our method achieves the smallest model size of 19MB as well as being competitive at 50 FPS detection speed compared with general object detectors. 
The source code is available at \url{https://github.com/MVME-HBUT/Faster-OreFSDet}.

\end{abstract}
 



\maketitle

\input{files/1-Introduction}
\input{files/2-Related_works}
\input{files/3-Methods}

\input{files/4-Experiments}
\input{files/5-Conclusion}

\noindent{\textbf{CRediT authorship contribution statement}}
\\
\\
\textbf{Yang Zhang}: Conceptualization, Methodology, Formal analysis, Data curation, Resources, Writing - original draft, Writing - review \& editing. \textbf{Le~Cheng}: Writing - original draft, Writing - review \& editing, Software, Data curation, Formal analysis, Validation, Visualization. \textbf{Yuting~Peng}: Writing - original draft, Data curation. \textbf{Chengming~Xu}: Writing - original draft \& review \& editing, Formal analysis, Data curation. \textbf{Yanwei~Fu}: Writing - review \& editing, Formal analysis, Data curation. \textbf{Bo~Wu}: Writing - review \& editing, Formal analysis, Data curation. \textbf{Guodong Sun}: Supervision, Funding acquisition, Writing - review \& editing
\\
\\
\noindent{\textbf{Acknowledgement}}
\\
\\
This work was supported in part by the National Natural Science Foundation of China (Grant 51775177), State Key Laboratory of Novel Software Technology (Grant KFKT2022B38), Hubei Key Laboratory of Modern Manufacturing Quality Engineering (Grant KFJJ-2022014), and the PhD early development program of Hubei University of Technology (Grant XJ2021003801).
\\
\\
\noindent{\textbf{Declaration of Competing Interest}}
\\
\\
The authors declare that they have no known competing financial interests or personal relationships that could have appeared to influence the work reported in this paper.

\printcredits

\bibliographystyle{cas-model2-names}

\bibliography{refs}

\bio{}
\endbio

\end{document}

%% file: files/1-Introduction.tex
\section{Introduction}
The particle size of the ore is an important data index to determine the crushing effect of the ore. Accurate and efficient detection of ore particle size is the basis of ore crushing optimization, which has a direct impact on the productivity of the entire beneficiation process.
Complex beneficiation site environments, dense adhesion, and stacking have posed great difficulties for ore particle size detection. In addition, some ores are reflected in the mineral processing workshop after using light, which makes it more difficult to distinguish the ore from the background.

Some scholars have proposed some traditional techniques \cite{ore_image} for detecting ores particle sizes.  
To achieve good performance, these approaches require laborious parameter adjustment processes. 
With the development of convolutional neural networks (CNN), there have been some significant advancements in object detection. 
However, general object detectors require a large amount of box labels to train, and obtaining such high-quality ore labeling data is expensive and time-consuming.
As labeled data become scarcer, the CNNs are easily overfitting and fail to be generalized.
Therefore, object detectors have trouble detecting real-world scenarios involving novel objects that are absent from common datasets for object detection.
On the basis of few-shot learning, some approaches \cite{SaberNet,CUI2022108296,Multigranularity} have developed insightful ideas for addressing data scarcity.

\begin{figure}[!t]
  \centerline{\includegraphics[width=5in]{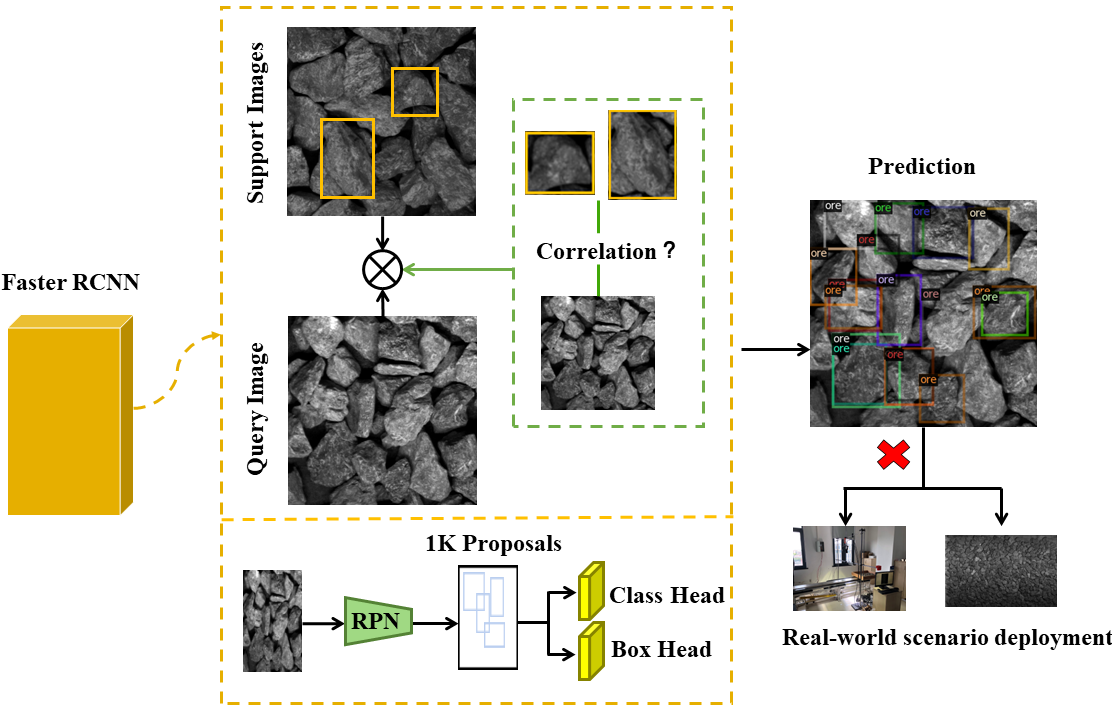}}
  \caption{Faster R-CNN is usually served as the basic detector of FSOD. This anchor-based few-shot object detector uses RPN to maximize the recall of the top 1K proposals and does not use these proposal scores in the test phase. A large number of proposals slows the speed.
  In addition, a large number of complex modules are designed to establish the correlation between support and query, 
  which leads to slow detection speed and high memory requirements. 
  It is extremely challenging for scenario deployments with limited computing resources and tight memory budgets.}
  \label{question}
\end{figure}

Few-shot object detection (FSOD) is the combination of traditional object detection and few-shot learning, which aims to predict and locate the object under a few annotated training samples.
As a result, it lessens the workload associated with labeling substantial volumes of data in the target domain.
However, the existing FSOD methods are mainly based on the traditional Faster RCNN \cite{Faster_RCNN}, as shown in Fig. \ref{question}. 
This two-stage detector includes a slow and independent region proposal generation step. In addition, to reduce the loss of accuracy caused by the lack of training data, a large number of complex modules are designed to establish the correlation between support
and query, which leads to slow detection speed and high memory usage. It is extremely challenging for scenario
deployments with limited computing resources and tight memory budgets.

To solve these problems, a well-known detector CenterNet2 \cite{centernet2} is served as the basic detector for the FSOD task. 
The CenterNet2 employs CenterNet with the ability to create accurate probability in the first stage, 
which is more accurate than the two-stage detector. 
Additionally, it enables the detector to employ fewer proposals (256 vs. 1K) in the region of interest (RoI) head, 
enhancing the overall accuracy and speed.
\par Furthermore, we design a support-feature mining block (SM Block) and relationship guidance block (RG Block) to fully establish the relationship between support and query features.
Specifically, adhesion, occlusion, and variations in ore appearance are particularly common, which present great difficulty in establishing high-quality support features.  
If sufficient discriminant information is not provided, the model can hardly learn the crucial features for class and bounding box predictions.
The SM Block is first suggested to encode feature representations along the height and width dimensions using linear projection.
The suggested SM block has the ability to assess the significance of the feature data provided by the ore images and remove detection interference brought on by the addition of background noise.
Next, we establish the spatial and channel correlation between support and query in the RG Block, which significantly improves the guidance performance of the query branch.
Finally, we propose a dual-scale semantic aggregation (DSA) module that retrieves detailed features at different resolutions for final classification and bounding box regression. Extensive experiments show the effectiveness of our method in comparison with state-of-the-art detectors.

Our main contributions can be summarized as follows:
\begin{enumerate}[(1)]
    \item A real-time few-shot object detector is designed for ore particle detection, which can alleviate the over-fitting issue when dealing with limited labeled data and significantly improve the performance of the FSOD task for the ore images. 
    \item We propose the SM Block to characterize the importance of semantic information in support features, and the RG Block to better establish the correlation between support and query features for guiding the generation of precise candidate proposals. 
    \item The proposed DSA module is designed to retrieve detailed features at different resolutions for final classification and bounding box regression.
\end{enumerate}

The remainder of this paper is structured as follows. In Section II, we provide a brief introduction to the ore image processing, general object detection, and FSOD methods. Section III presents the problem definition and four components. Section IV details experimental procedures and results over the ore dataset. Finally, the conclusion is drawn in Section V.

%% file: files/2-Related_works.tex
\section{RELATED WORK}
\subsection{Ore Image Processing} 
The particle size analysis task is usually aimed at the ores on the belt and can be divided into three modes: particle size statistics, particle size classification, and large block detection. 

Particle size statistics refers to the determined value of ore size in an image that is obtained. Generally, the semantic segmentation network is used to segment each object in the image, and then the number of pixels of each object is obtained by OpenCV and other toolkits. According to the relationship between the unit pixel and the actual size, the area $S$ of each ore is obtained. Finally, the corresponding ore particle size statistics are completed according to the actual needs, such as the particle size $d$ (the equivalent circle diameter corresponding to the area $S$) of each ore in the image. 
Some researchers have proposed some solutions and achieved good results.
The primary measures are regression-based classifiers and techniques based on certain theories \cite{ore_image}. 
Watershed transform processes \cite{particle} were introduced in region-based segmentation techniques for ore particle sizes.
However, it is difficult to adapt these methods to different situations since they require a time-consuming parameter change procedure to achieve satisfactory performance. The development of CNN-based image classification has led to significant advancements in downstream fields such as object detection and semantic segmentation. 
For example, Liu et al. \cite{LiuXiaobo2020OreIS} used U-Net to detect ore particle size, and Li et al. \cite{UNet} also proposed a U-Net-based model that alleviated ore particle size detection issues by improving the loss function and utilizing the watershed technique.
Liu et al. \cite{oresegmentation} used morphological transformation to process the mineral image mask and segment the key areas in the mineral image.
Sun et al. \cite{Efficient} proposed a novel efficient and lightweight method for ore particle size detection. 
Particle size classification refers to the classification of particle size grades by considering all ores in the image as a whole. In general, the ore datasets of different particle sizes are constructed first, and then the image classification network is trained. Finally, the trained model is used to classify the unknown ore images. Olivier et al. \cite{Olivier2020EstimatingOP} used VGG16 network~\cite{VGG} to classify 10 particle size grades of ore images, which provided guidance for subsequent mine production operation control.
Large block detection refers to the identification of oversized ores on the belt. The object detection network is first used to obtain the coordinate information of the ore, then the external rectangular area of each ore is calculated. Finally, it will compare with the set threshold to determine whether there is a large block on the belt. At present, there are few related researches, and this paper is a study of the third detection task. 
\subsection{General Object Detection}
Object detection is to identify and categorize a variety of targets in the images and then determine the categories of numerous objects as well as their locations. 
Object detection has generally been the most challenging subject in the field of computer vision since different things might have a wide range of appearances, shapes, and attitudes.
The deep learning framework may be utilized with one-stage and two-stage mainstream object detectors.
The former, such as the well-known Faster R-CNN \cite{Faster_RCNN}, first created category-unknown region proposals in the form of RPN, 
and then projected the proposals onto the feature maps following the RoI pooling.
Finally, proposal features were fed into the fully connected layer for classification and regression to determine class labels and fine-tune bounding boxes.
Grid RCNN \cite{Grid_R_CNN} introduced a grid-guided localization mechanism for accurate object detection. 
Cascade RCNN \cite{Cascade_R_CNN} used cascade regression as a resampling mechanism to improve the performance of the probe by increasing the intersection over union (IoU) of the proposal stage by stage.
You only look once (YOLO) series \cite{yolov3,YOLOf} and single shot multi-box detector (SSD) \cite{SSD} were examples of one-stage detectors that provided a non-region proposal framework for class and bounding box prediction. 

On the other hand, there are two categories of existing object detection: anchor-based and anchor-free.
Faster R-CNN \cite{Faster_RCNN} was the one that initially put out the idea of anchors. 
A proper initialization for RPN allows it to avoid using an unnecessary amount of search space and produce better region suggestions since each anchor represents a predetermined box with a certain aspect ratio and size.
Many one-stage detectors also employ anchors to raise the quality of their proposals. 
However, anchors add a lot of hyper-parameters and the imbalance between positive and negative proposals is exacerbated since most anchors do not include targets.
Numerous anchor-free techniques were then proposed, such as FCOS~\cite{FCOS}, which directly predicted using the center-based technique and 
positive and negative samples are defined using various methods.
In addition, there are some methods designed to solve special problems in object detection.
RetinaNet \cite{RetinaNet} proposed focal loss to alleviate the problem of foreground-background class imbalance. 
Based on focal Loss, VarifocalNet \cite{VarifocalNet} used Varifocal Loss to predict each image for dense object detection.
Li et al. \cite{gfl} proposed a generalized focal loss via joint quality estimation and classification.
Shuang et al. \cite{Scalebalanced} proposed a loss function to alleviate a matching imbalance due to different scales of objects.
CenterNet2 \cite{centernet2} replaced RPN with CenterNet with the ability to generate accurate likelihood in the first phase, making it more accurate and faster.
According to the peculiarities of the parallel implementation of classification and localization in conventional one-stage object detection, 
Probabilistic anchor assignment (PAA) \cite{paa} proposed a new anchor assignment strategy, which adaptively assigns labels to anchors in the form of probability according to the training state of the model.


\subsection{Few-shot Object Detection}
With only a few training samples, the FSOD task attempts to tackle the object detection issue. 
There are two main categories: transfer learning-based and meta-learning-based methods.
For transfer learning-based methods \cite{MPSR}, the target domain model was first initialized using the model parameters from the source domain model on large-scale datasets, and then fine-tuned on small-scale datasets.
There are two main methods of meta-learning-based approaches \cite{MetaLearning}: learn to fine-tune and learn to measure. 
The former is to learn category-agnostic parameters for new categories and specific weights on new tasks. 
Two-stage fine-tuning approach (TFA) \cite{TFA} first trained the entire detector on a data-rich base class, and then fine-tuned the last layer of the detector on a small balanced training set consisting of base classes and new classes while freezing other parameters of the model.
Sun et al.\cite{FSCE} provided a comparative learning method into the two-stage fine-tuning method to reduce the intra-class differences and increased the inter-class differences.
In contrast, learning to measure requires the feature fusion of the query set and the support set to complete an example search in a constrained number of support sets.
However, how these features are incorporated, where they are integrated, and what training strategies are employed vary depending on the model.
Features in two-stage methods such as Meta RCNN \cite{Meta_RCNN} and FsDetView \cite{FSDetView} are fused after the RPN. 
Similarly, the feature fusion of Meta YOLO \cite{METAYOLO} was directly  performed before the detection head. 
AttentionRPN \cite{AttentionRPN} provided a feature fusion module to exclude proposals by category information.
In addition, there are some other methods to deal with the special problems in this field. 
Wu et al.~\cite{MPSR} proposed a multi-scale positive sample refinement (MPSR) method to enrich the object scale in FSOD.
A dual attention strategy was introduced by dual-awareness attention for few-shot object detection (DAnA) \cite{Dual_Awareness} to address the issues of spatial information loss brought on by global pooling and spatial imbalance brought on by convolutional attention.
Spatial reasoning is introduced into few-shot object detection to detect new categories in \cite{Spatial}, which is a novel approach in this field.

In our framework, we use the AttentionRPN \cite{AttentionRPN} as the baseline and further improve the performance on ore images. 
Fine-tuning FSOD method is adopted with CenterNet2 \cite{centernet2} as the basic detection framework. 
Different from the previous works, we develop a lightweight and effective few-shot object detector on CenterNet2 \cite{centernet2}.  
Compared to the two-stage anchor-based detector, our method uses CenterNet with the ability to generate accurate likelihood, 
which is more accurate and allows the detector to use fewer proposals (256 vs 1K) on the RoI head.
In the second stage, we adopt a uniquely designed lightweight detection head for ore images, making our detector more accurate and faster overall.
In addition, we design lightweight and effective few-shot strategies to generate higher quality supports and establish the effective correlation between support and query features for more precise guidance.

%% file: files/3-Methods.tex
\section{METHODOLOGY}
In this section, we first introduce the motivation overview. Subsequently, we introduce SM block to characterize the importance of semantic information in support features and RG Block to better establish the correlation between support and query features for guiding the generation of precise candidate proposals. Finally, a dual-scale semantic aggregation module is designed to retrieve detailed features at different resolutions for final classification and bounding box regression.

\subsection{Motivation Overview}
\begin{figure*}[!t]
  \centering
  \includegraphics[width=6.2in]{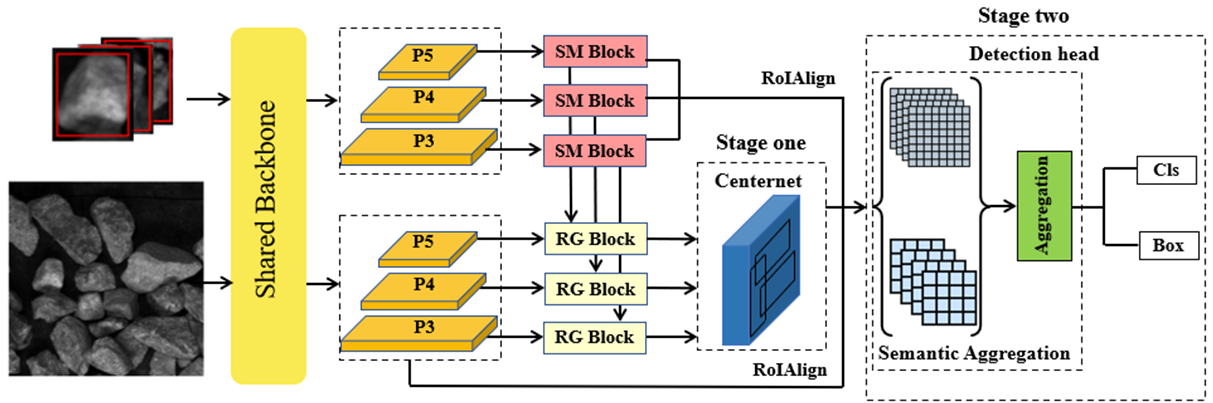}
  \caption{The overall framework of our proposed method. 
  The training input of each episode consists of a query image and several support images from one class. 
  The shared feature extractor and feature pyramid networks (FPN) first extract the query and support features.
  The P3, P4, and P5 of support features from FPN are fed into the SM Block for effective feature representation and then fed into the RG Block to generate attention maps with the same level query features from FPN. These attention feature maps are sent to the one-stage detector CenterNet. 
  After filtering out the negative objects that do not belong to the support category, accurate candidate proposals are generated. 
  Subsequently, the candidate proposals and the support features are sent to the detection head, where a more accurate parsing is conducted between the support box and the potential box in the query feature maps.
}
  \label{Framework}
\end{figure*}

It is a challenging task to realize the mobile deployment of an ore detection model with limited training data in a complex detection environment.
A lack of training data will cause general object detectors to be overfitting.
Despite alleviating the overfitting issue, FSOD methods usually suffer some difficulties caused by the poor performance of speed and accuracy, and excessive model size, especially when used to embedded mobile devices.
Specifically, there are two main reasons for the poor performance of the existing FSOD methods:
(i) There is just one class of ore in the ore particle size detection, only a straightforward classification of the foreground and background is required. However, the previous works generally employed the two-stage Faster RCNN \cite{Faster_RCNN} as the detection framework. Faster RCNN generated a series of anchor boxes at each anchor according to certain rules, and then adjusted proposals (RoIs) at the second stage for anchors combined with neural network output bias and a series of selection rules. When the existing methods are directly migrated to the ore particle size detection, this detector exhibits slow speed and redundant classification full connection layer. In addition, there are a large number of complex modules to decrease the accuracy loss caused by limited training data, which resulted in slow detection speeds and enormous memory requirements.
(ii) The occlusion and adhesion of ore appearance cause incomplete expression of characteristics. Therefore, when establishing the correlation between support and query features for guidance, the support feature information of the ore is particularly important. Under few-shot task setting, however, the ore class prototype is derived from the characteristics of the global average pooled support feature maps, which results in the loss of particular local contexts. The correlation between ore image support and query features cannot thus be clearly established.

As shown in Fig. \ref{Framework}, we present a lightweight and faster few-shot detector on CenterNet2 \cite{centernet2}.
The proposed SM Block characterizes the importance of location information carried by support features, and RG Block makes full use of support features to guide the generation of accurate candidate proposals. 
The DSA retrieves detailed features at different resolutions to contribute with the prediction process.

\begin{figure}[!t]    
  \centering
  \includegraphics[width=4.5in]{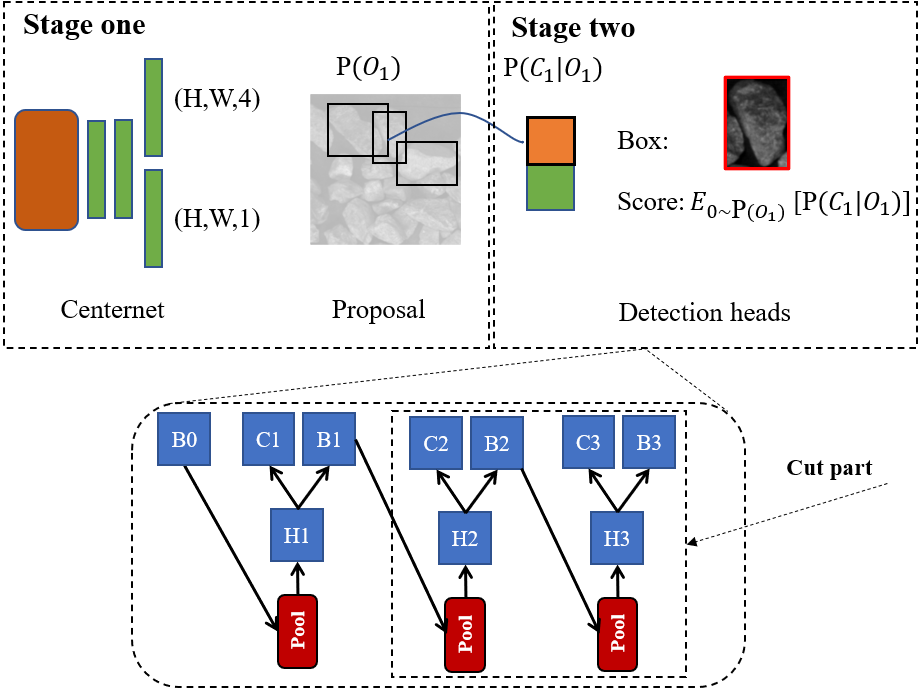}   
  \caption{ The structure of CenterNet2.
  A lightweight VoVNet \cite{vovnet} is served as backbone. 
  To extract and classify region-level characteristics in stage one, CenterNet2 \cite{centernet2} employs the CenterNet with the ability to create precise probability. 
  In stage two, we remove the extra two heads and shrink the number of head channels to a smaller 128. To increase the log-likelihood of GT targets, these two stages are trained concurrently. The final log-likelihood is used by our detector as the detection score in inference.}
  \label{centernet2}
\end{figure}

\subsection{A Faster and Lighter Detection Framework}
Most FSOD methods are built on a two-stage detector Faster RCNN \cite{Faster_RCNN}.
All two-stage detectors used a weak RPN to maximize the recall of the first 1K proposals and did not utilize these proposal scores during the test phase. 
A large number of proposals slows the speed, and the proposal network based on recall considerations does not directly provide a clear probability explanation like the one-stage detector. 
Additionally, the last fully connected layer used by the original Faster RCNN \cite{Faster_RCNN} takes up a large portion of the parameters. 
All RoIs after RoI pooling will go through this full connection and are calculated separately without shared computing. 

To obtain a lighter and stronger detection framework, our method is built on the two-stage detector CenterNet2 \cite{centernet2}, as shown in Fig. \ref{centernet2}. 
Compared to the two-stage anchor-based detector, CenterNet2 \cite{centernet2} uses the CenterNet with the ability to generate accurate likelihood in the first stage, which is more accurate and allows the detector to use fewer proposals (256 vs 1K) on the RoI head, making our detector more accurate and faster overall. 
In addition, CenterNet2 \cite{centernet2} employs a complex cascade RCNN architecture in the second stage, which we customize to achieve a lightweight design due to the single class characteristics of ore images.

\begin{figure}[!t]    
  \centering
  \includegraphics[width=6.2in]{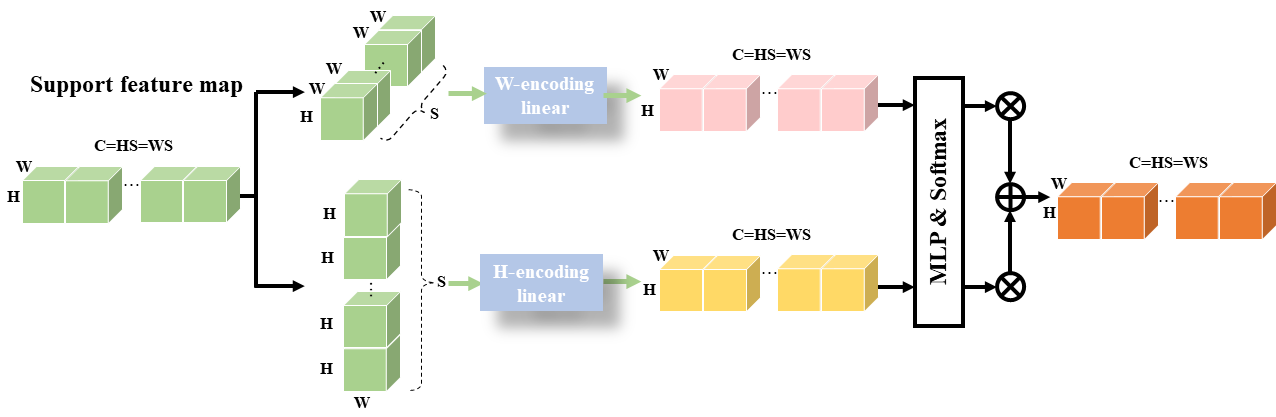} 
  \caption{ The structure of SM block.
  Two branches in SM Block are in charge of encoding information along the height and width dimensions. 
  The dimension information supplied by the two branches is adaptively aggregated using the attention method to acquire the position-sensitive information of the final support.
  }
  \label{SM}
\end{figure}
 
\subsection{Support Feature Mining Block}
\textbf{Feature information Mining.} 
The quality of supports is crucial in determining how to guide the query branch.  
In previous work, the supports from the backbone were frequently used directly, which introduced distracting background noise. 
To this end, we propose a simple and data-efficient SM Block that characterizes the importance of the location information carried by supports. 

In Fig. \ref{SM}, our module consists of two branches that are responsible for encoding information along the height and width dimensions.
When encoding spatial information along the height dimension, the height channel permutation operation is performed first.
Given $\bm{X}\in t^{ H\times W\times C}$, to satisfy $C = H * S$, we first divide it into $S$ parts along the channel dimension,
and then perform a high-channel permutation operation to get $[H_1,H_2 \cdots H_s]$.
Next, the height information is encoded through a fully connected layer, followed by a height channel permutation operation to restore the dimension information to $\bm{X}$.
Similarly, these operations are performed in the width direction. 
Finally, the weighted sum of two branches ($K=2$) is carried out, which is described as follows:

Multilayer Perceptron (MLP) for feature mapping of the two summed branches
$$\bm{z}=T_{MLP}(((\bm{X_h+X_w})\bm{R_1})\bm{R_2}), \quad        \bm{R_1}\in \mathbb{R}^{ C\times \hat{C}},\bm{R_2}\in \mathbb{R}^{ \hat{C}\times KC}. \eqno(1) $$ 

$Reshape$ and $softmax$
$$\bm{z}\rightarrow\bm{Z}\in \mathbb{R}^{ K\times C}. \eqno(2) $$  

Weighted summation
$$\bm{\hat{X}}=\sum_{k=1}^K\bm{X_k}[i,:]\odot \bm{Z}[k,:]. \eqno(3) $$ 

The SM block can capture long-range dependencies along one spatial direction while retaining accurate location information in the other direction. 
The output features of the location-aware obtained in this way are aggregated in a complementary way to form an effective representation of the target of interest.

\subsection{Relationship Guidance Block}

\begin{figure*}[!t]
  \centering
  \includegraphics[width=3.5in]{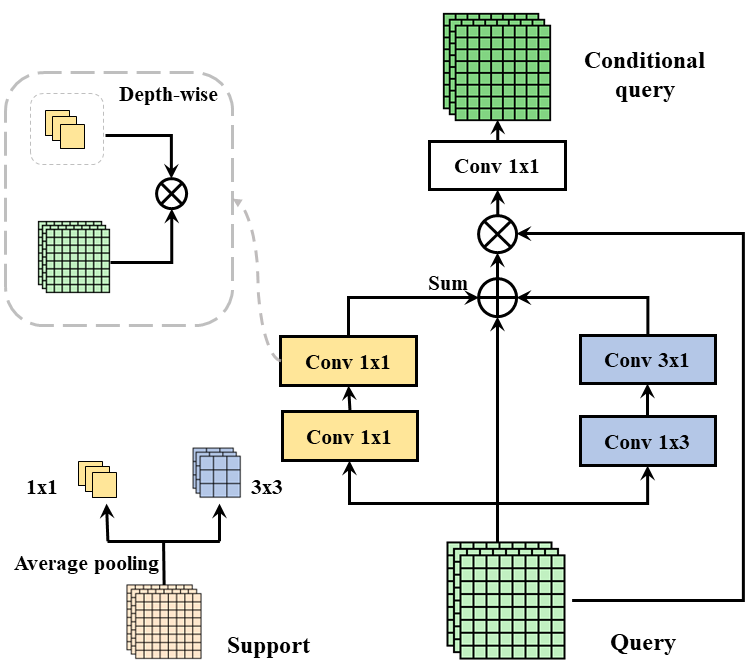}
  \caption{The structure of RG Block. 
  Support is pooled into two different sizes of kernels with ore prototype spatial information, which are then convoluted channel by channel on the query map. 
  The output two feature maps are then superimposed on the original query map to get the final attention map with spatial scale correlation. 
  Finally, the query map and attention map are concatenated along the channel to establish the feature channel correlation.
}
  \label{RG}  
\end{figure*} 

After obtaining a high-quality support class prototype, an effective relationship between query and support is crucial to the performance of the model. Previous work AttentionRPN\cite{AttentionRPN} performed a global average pooling operation on support 
and used it as a convolution kernel $(1\times 1)$ to slide over the query feature map to obtain an attention map with support spatial information. 
However, when employing the same technique on ore images, this global average pooling operation results in the loss of support 
and only concentrates on spatial information, whereas channel information relating to categories is not correlated. 
Therefore, we suggest a RG Block to fully build an effective relationship between support and query as indicated in Fig. \ref{RG}.

\textbf{Spatial scale correlation}: the category of the target is closely related to the appearance, which is determined by the feature's spatial dimension. 
Consequently, the spatial correlation between the two features might substantially indicate how similar they are to one another. 
To retain more support spatial information for query context guidance, we pool supports into 1x1, and 3x3 sizes, and perform parallel convolution operations on the query. 
The 3x3 convolution is divided into 1x3 and 3x1 deep strip convolutions to further decrease the computational cost 
while facilitating the extraction of banded ore characteristics.
As follows, we specify the spatial scale correlation $\mathcal{R}_{\mathrm{s}}$
$$\mathcal{R}_{\mathrm{s}}(\mathbf{X}, \mathbf{Y})=\bm{Q}_{c,h,w}=\sum_m^P\sum_n^P\bm{X}_{c, m, n}\cdot \bm{Y}_{c, h+m-1, w+n-1}, \eqno(4)$$where $\bm{Q}$ represent the generated attention feature maps. Each support feature map of $\bm{X}\in t^{ P\times P\times C}$ ($P$ denotes the kernel size after pooling) is served as a convolution kernel for convolution operations on the corresponding query feature map of $\bm{Y}$ in depth cross-correlation.
After spatial scale correlation, we superimpose two feature maps with different spatial information of supports onto the original query feature maps to get the final feature maps.

\textbf{Feature channel correlation}: prior research has demonstrated that the category information of images is frequently present in the feature channel. 
Along the distribution of the channel, the deep features of the same category are similar. 
We use the following criteria to define the similarity $\mathcal{R}_{\mathrm{c}}$
$$
\mathcal{R}_{\mathrm{c}}(\mathbf{X}, \mathbf{Y})=\operatorname{Conv}(\operatorname{Cat}(\mathbf{X}, \mathbf{Y})), \eqno(5)
$$where Cat represents two features concatenated along the channel, and the interaction between the channels is modeled using a normal $1 \times 1$ convolution.

\subsection{Dual-scale Semantic Aggregation Module}
\begin{figure}[!t]    
  \centering
  \includegraphics[width=4.3in]{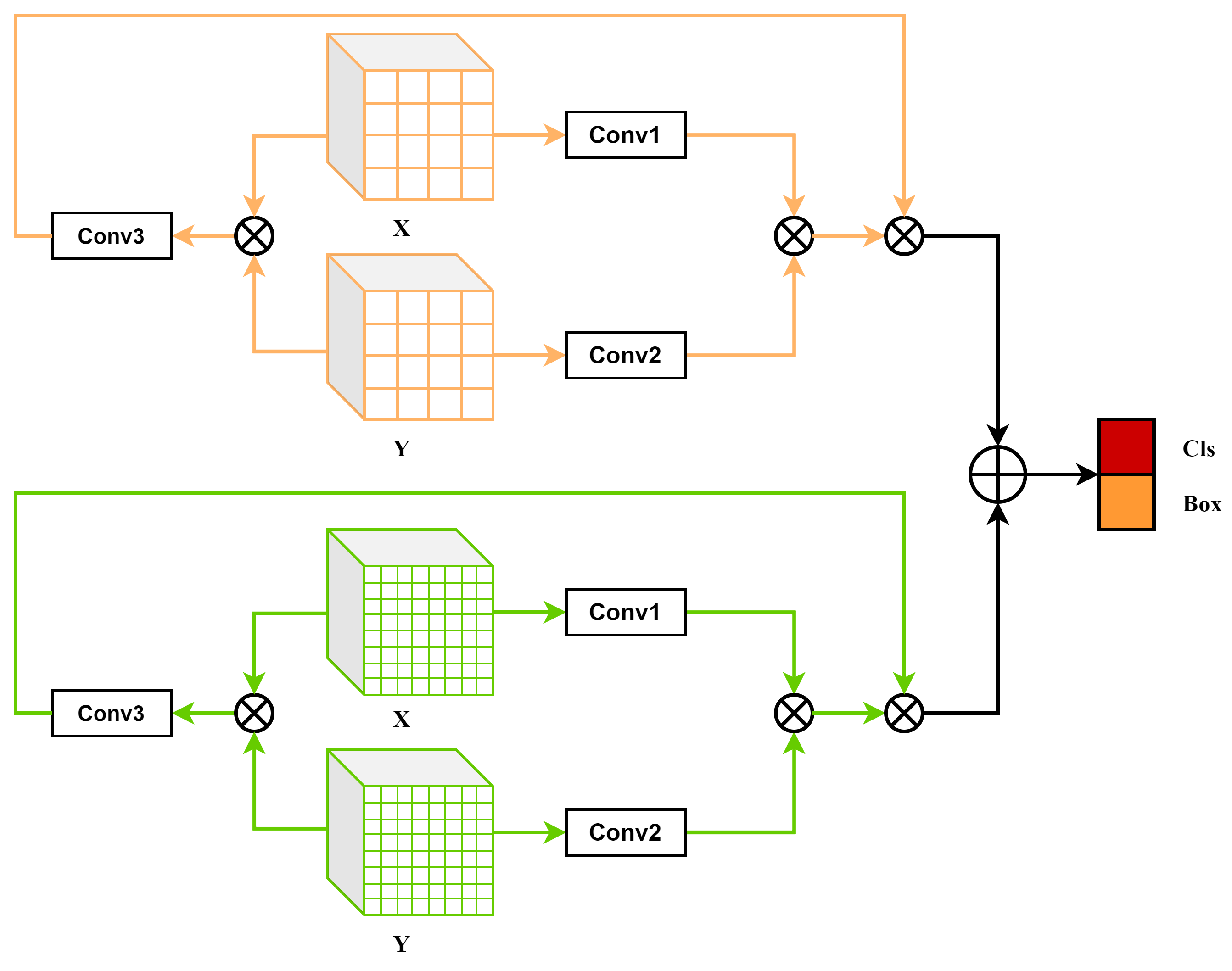}
  \caption{ A global matching relationship is established between support features $\bm{X}$  and query features $\bm{Y}$.
  Then, two different resolution feature maps are aggregated to provide a more comprehensive feature representation for final classification and box regression.
  }
  \label{dual-scale Semantic Aggregation}
\end{figure}

After one stage, the RoI align module performs feature extraction for final class prediction and bounding box regression. 
Based on previous experience, implementation with a fixed resolution of 8 may cause information loss during training.
An abundance of training data can make up for this information loss in general object detection, while the issue gets severe for few-shot object detection with few shots. 
Therefore, we propose a dual-scale semantic aggregation module as shown in Fig. \ref{dual-scale Semantic Aggregation}. 

Empirically speaking, small resolution tends to focus on large target information, 
while larger resolution tends to focus on smaller object information. 
Since the ore image sizes are only medium and large, we choose 4 and 8 resolutions and perform parallel pooling. 
In addition, to further guide more accurate classification and bounding box regression using support information in the second stage, 
we establish a global matching relationship between the support feature map and the query feature map
$$
\mathcal{A}_{\mathrm{DSA}}(\mathbf{X}, \mathbf{Y})=\operatorname{Conv3}(\operatorname{Cat}(\mathbf{X}, \mathbf{Y}))+\operatorname{Cat}(\operatorname{Conv1}(\mathbf{X}), \operatorname{Conv2}(\mathbf{Y})). \eqno(6)
$$
\par Finally, we aggregate two different resolution feature maps to obtain a more comprehensive feature representation for final classification and bounding box regression.

%% file: files/4-Experiments.tex
\section{EXPERIMENTS}
In this section, we first introduce the implementation details, datasets, and evaluation metrics. Then, we conduct ablation for a lightweight few-shot detector and evaluate the effectiveness of SM Block, RG block, and DSA by comprehensive experiments.
Finally, we compare our proposed OreFSDet with the state-of-the-art methods on the ore dataset. 

\subsection{Experiments Setup}
\subsubsection{Implementation Details}
The CenterNet2 \cite{centernet2} serves as the detection framework of our network throughout the experiments.
During the training process, for the constructed probabilistic interpretation of two-stage detection, the loss calculation in CenterNet is used in the first stage, and the loss in Cascade RCNN is employed in the second stage. Finally, the above two losses are added to obtain the total loss. 
We use a fine-tuning method to achieve detection under a few-shot scenario.
The 80 categories in Microsoft COCO 2017 \cite{coco} are employed as base classes for base training. 
And in place of the fully-connected layer, a newly created layer with a random initialization value is created for new class ore.
A significant aspect of our learning process is freezing the backbone and incorporating the newly designed network model in the second fine-tuning phase.
For fine-tuning on the ore image dataset, the number of iterations is set to 20000 and the learning rate is 0.001.
The batch size is set to 1 with a single NVIDIA GTX2080Ti GPU. 
The scaled query images have a short side of 320 pixels and a long side of fewer than 1000 pixels. 
Support images are $240 \times 240$ pixels with zero-padded.

\subsubsection{Datasets}

\begin{figure}[!t]
	\centering 
	{\begin{minipage}[t]{2.2in}
        \centering
        {\includegraphics[height=6cm]{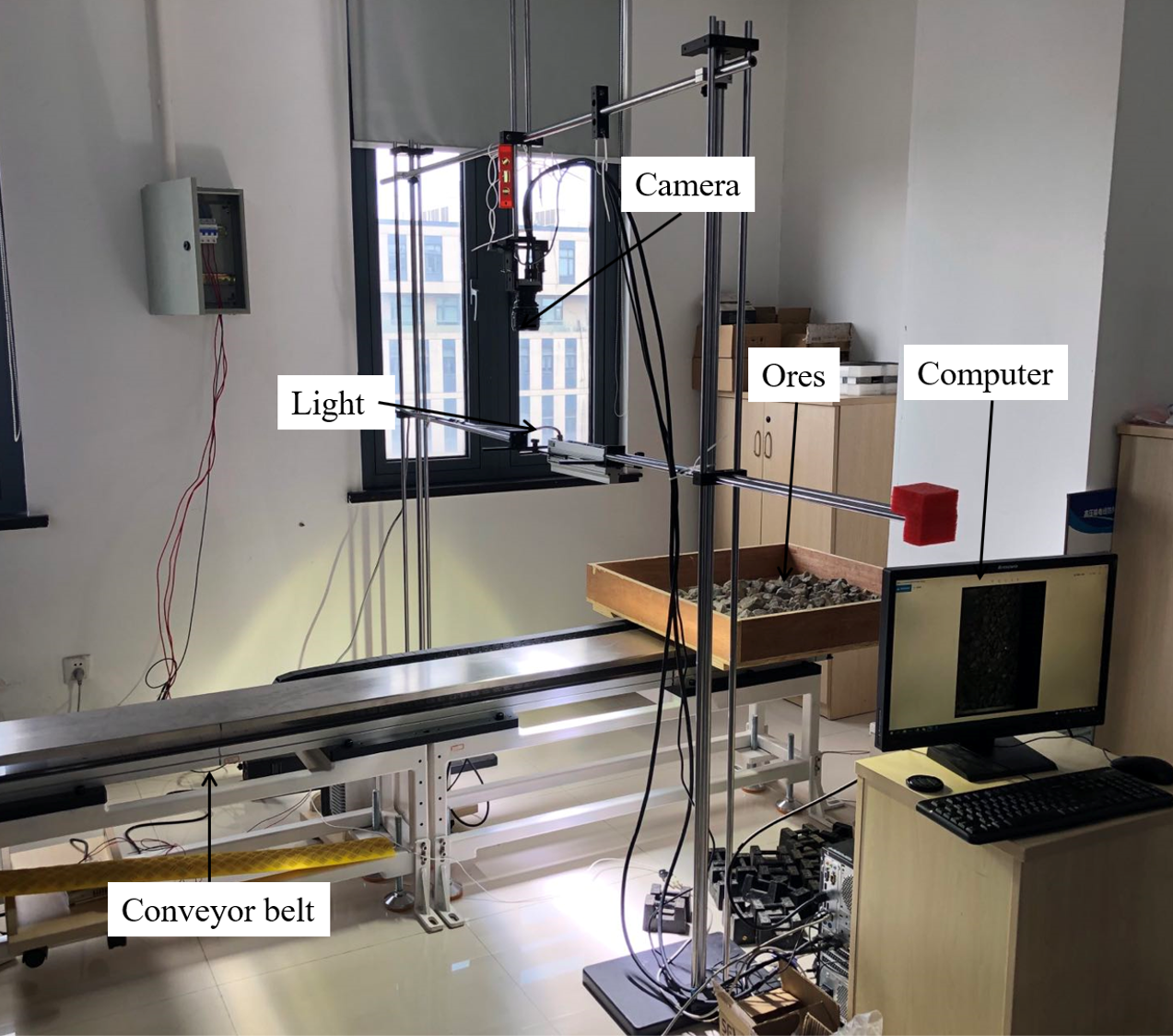}}
		\centerline{Experimental Platform}
	\end{minipage}
        \vspace{3pt}
	\begin{minipage}[t]{2.2in}   
	\centering
        {\includegraphics[height=6cm]{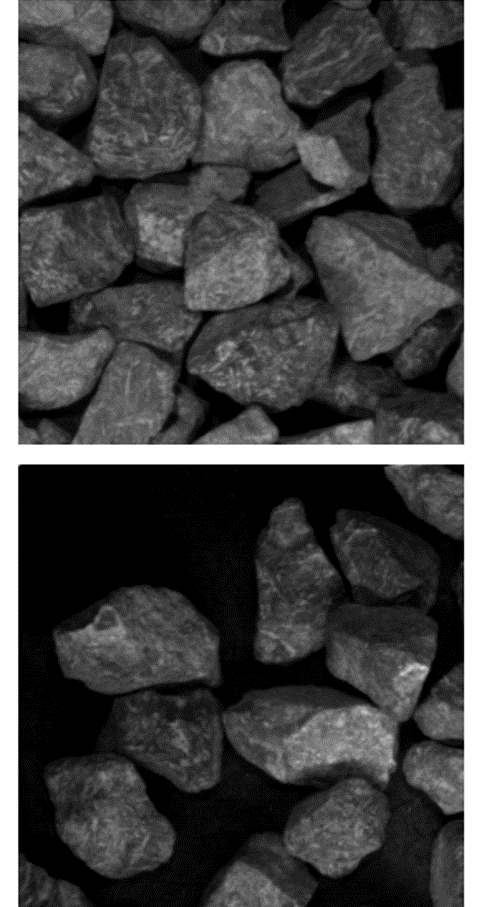}}
	 
		\centerline{Ore Images}
	\end{minipage}}
	\caption{We collect ore images with different densities and build an experimental platform for ore image detection. }
	\label{experiment pictures}
\end{figure}

The MS COCO~\cite{coco} is set up for base training and ore image dataset for fine-tuning training. 
The MS COCO is a large-scale dataset for image detection, semantic segmentation, and other vision tasks. 
It includes 80 target categories, 1.5 million targets, and more than 330K photos.
Figure \ref{experiment pictures} shows how the experiment platform is used to collect ore images of different sizes for the ore image dataset. 
The ore distribution is then changed on different scales by making it dense, sparse, etc. 
In order to be used in network training, we also split huge ore images into smaller ones.

\subsubsection{Evaluation Metrics}
A total of eight indicators are utilized to determine whether the proposed method is effective:
${AP^{box}}$, ${AP^{box}_{50}}$, ${AP^{box}_{75}}$, $AP^{m}$, $AP^{l}$, frame per second (FPS), inference memory consumption, and model size.
The accuracy evaluation metrics adopt the COCO\cite{coco} evaluation metrics commonly used in object detection(top five). 
${AP^{box}}$(average precision of boxes) represents the average of precision of boxes under different box IOU values. IOU values vary from 0.5 to 0.95, with a calculation of precision of boxes at an interval of 0.05. ${AP^{box}_{50}}$ and ${AP^{box}_{75}}$ represent values of precision of boxes with IOU of 0.5 and 0.75, respectively. The $AP^{s}$, $AP^{m}$ and $AP^{l}$ represent the three categories of small, medium and large. The pixels with an area less than $32\times32$ are classified as small targets, those with an area greater than $96\times96$ are classified as large targets, and those between the two are medium targets. There are only medium and large ore targets.
The frame per second (FPS) is used to indicate how much a model costs to compute. 
Memory consumption during inference demonstrates the model's dependence on the hardware.
The last factor is critical in determining whether or not a model can be implemented on low-cost hardware.
It is crucial to consider the last factor when determining whether or not a model can be implemented on low-cost hardware. 
To further validate the effectiveness, we train 4060 images for general object detection in experiments. 
In contrast, the shot is 10 for our proposed OreFSDet, and the validation set is the same for both of them at 1060 images.

\subsection{Ablation Study}

\begin{table}[!t]
	\renewcommand{\arraystretch}{1.5}
	\centering
	 \caption{The influence of the number of heads and corresponding channels in CenterNet2 on the lightweight design of the model}
	
	  \begin{tabular}{cccccccccc}
	  \toprule
	 
	  \multicolumn{2}{c}{\multirow{1.8}[2]{*}{\textbf{Head\_channel}}} & \multicolumn{3}{c}{\textbf{Cascade\_head}} & \multirow{1.8}[2]{*}{\textbf{$\bm{AP^{box}}$}} & \multirow{1.8}[2]{*}{\textbf{$\bm{AP^{box}_{50}}$}} & \multirow{1.8}[2]{*}{\textbf{$\bm{AP^{box}_{75}}$}} & \multirow{1.8}[2]{*}{\textbf{FPS}} & \multirow{1.8}[2]{*}{\makecell[c]{\textbf{Model Size}\\\textbf{(MB)}}} \\
	  \cmidrule{3-5}    
	  \multicolumn{2}{c}{} & \textbf{stage1(0.6)} & \textbf{stage2(0.7)} & \textbf{stage3(0.8)} &       &       &       &       &  \\
	   \midrule
	  \multicolumn{2}{c}{    } & \CheckmarkBold     &       &       & 53.0 & \textbf{77.6} & 63.3 & \textbf{43} & \textbf{62.3} \\
	  \multicolumn{2}{c}{1024} & \CheckmarkBold     & \CheckmarkBold     &       & \textbf{54.1} & 76.3 & \textbf{65.0} & 40 & 102.0 \\
	  \multicolumn{2}{c}{    } & \CheckmarkBold    & \CheckmarkBold     & \CheckmarkBold     & 53.0 & 74.2 & 63.2 & 33  & 138.0 \\
	  \midrule
	  \multicolumn{2}{c}{   } & \CheckmarkBold     &       &       & 52.4 &\textbf{77.6} & 62.7 & \textbf{42} & \textbf{36.2} \\
	  \multicolumn{2}{c}{512} & \CheckmarkBold     & \CheckmarkBold     &       & \textbf{54.2} & 76.4 & \textbf{65.4} & \textbf{42} & 54.3 \\
	  \multicolumn{2}{c}{   } & \CheckmarkBold     & \CheckmarkBold     & \CheckmarkBold     & 53.2 & 74.3 & 63.6 & 33  & 71.3 \\
	   \midrule
	  \multicolumn{2}{c}{   } & \CheckmarkBold     &       &       & 52.3 & \textbf{77.3} & 62.6 & \textbf{43} & \textbf{24.7} \\
	  \multicolumn{2}{c}{256} & \CheckmarkBold     & \CheckmarkBold     &       & \textbf{54.7} & 77.0 &\textbf{65.9}  & 37 & 33.2 \\
	  \multicolumn{2}{c}{   } & \CheckmarkBold     & \CheckmarkBold     & \CheckmarkBold     & 53.5 & 74.9 & 64.2 & 36 & 41.5 \\
	   \midrule
	 \multicolumn{2}{c}{   } & \CheckmarkBold     &       &       & 52.5 & \textbf{77.9} & 63.0 &\textbf{50}   & \textbf{19.4} \\
  \multicolumn{2}{c}{128} & \CheckmarkBold     & \CheckmarkBold     &       & \textbf{54.4} & 76.8 & \textbf{65.8} & 40 & 23.5 \\
	  \multicolumn{2}{c}{   } & \CheckmarkBold     & \CheckmarkBold     & \CheckmarkBold    & 52.9 & 74.7 & 63.3 & 33  & 27.6 \\
	  \bottomrule
	  \end{tabular}%
	\label{lightweight detector1}%
\end{table}%

\begin{table}[!t]
	\renewcommand{\arraystretch}{1.5}
	\centering
	\caption{The impact of backbone and few-shot strategies on the lightweight design of the model}
		\centering
	  \begin{tabular}{cccccccccc}
	  \toprule
  \multicolumn{3}{c}{\textbf{Module}} & \multirow{2}[2]{*}{\textbf{Framework}} & \multirow{2}[2]{*}{\textbf{Backbone}} & \multirow{2}[2]{*}{\textbf{$\bm{AP^{box}}$}} & \multirow{2}[2]{*}{\textbf{$\bm{AP^{box}_{50}}$}} & \multirow{2}[2]{*}{\textbf{$\bm{AP^{box}_{75}}$}} & \multirow{2}[2]{*}{\textbf{FPS}} & \multirow{2}[2]{*}{\makecell[c]{\textbf{Model Size}\\\textbf{(MB)}}} \\
  \cmidrule{1-3}    \textbf{SM Block} & \textbf{RG Block} & \textbf{   DSA   }   &       &       &       &       &       &       &  \\
	  \midrule
	 \multicolumn{1}{c}{\CheckmarkBold }     & \multicolumn{1}{c}{\CheckmarkBold }     & \multicolumn{1}{c}{\CheckmarkBold }     & CenterNet2 & ResNet50 \cite{Deep_Residual} & \textbf{52.9} & \textbf{78.4} & \textbf{64.0} & 38 & 130.0 \\[2 ex]
	  \multicolumn{1}{c}{\CheckmarkBold }     & \multicolumn{1}{c}{\CheckmarkBold }     & \multicolumn{1}{c}{\CheckmarkBold }     & CenterNet2 & DLA \cite{dla} & 48.9 & 65.3 & 58.2 & 29 &81.4 \\[2 ex]
	  \multicolumn{1}{c}{\CheckmarkBold }     & \multicolumn{1}{c}{\CheckmarkBold }     &       & CenterNet2 & VoVNet \cite{vovnet} & 51.2 & 76.1 & 61.2 & 51 & 16.5 \\[2 ex]
	  \multicolumn{1}{c}{\CheckmarkBold }     &       & \multicolumn{1}{c}{\CheckmarkBold }     & CenterNet2 & VoVNet \cite{vovnet} & 51.4 & 77.1 & 61.3 & 50  &19.4 \\[2 ex]
      \multicolumn{1}{c}{\CheckmarkBold }     & \multicolumn{1}{c}{\CheckmarkBold }     & \multicolumn{1}{c}{\CheckmarkBold }     & CenterNet2 & VoVNet \cite{vovnet} & 52.5 & 77.9 & 63.0 & \textbf{50}  & \textbf{19.4} \\[0.5 ex]
	  \bottomrule
	  \end{tabular}%
	\label{lightweight detector2}%
\end{table}%

\subsubsection{Ablation for a lightweight few-shot detector}
\par The three cascaded heads exist in the second stage detection of original CenterNet2 \cite{centernet2}. 
To find out which heads and corresponding IoU values are crucial for ore detection and the influence of the number of channels on the performance of the model, we conducted comprehensive experiments. 
It can be concluded from the table that as the number of channels in the head decreases sharply, 
only slight accuracy damage is caused, which is attributed to the single category detection and simple feature information of the ores. 
Compared with the single number of head, the multiple number of cascaded detection head generally brings accuracy improvement under different channels, but it is followed by the unbearable model size and detection speed burden. 
Therefore, we choose channel 128 and cut the three heads into one head, and the corresponding IoU value is 0.6.


To further achieve a smaller and speedier model, we chose a lighter backbone, as seen in Table \ref{lightweight detector2}. 
The lightweight VoVNet \cite{vovnet} significantly contributes reduced computational overhead and model size occupancy to our model compared to the backbone ResNet50 \cite{Deep_Residual} and DLA \cite{dla}.
Compared with the baseline, we are more than $7\times$ smaller in model size and nearly 19 frames faster in inference speed, 
which is of great significance for our model deployment on the edge.
As shown in Table \ref{lightweight detector2}, we verify the effectiveness of the three designed modules with lightweight CenterNet2 \cite{centernet2} as the baseline. 
As a module that processes support information separately, the SM Block needs to be used in conjunction with the other two modules. 
The combination of RG Block and DSA with SM Block improves the overall performance of the model.

\begin{table}[!t]
	\renewcommand{\arraystretch}{1.3}
	\centering
	\caption{The effectiveness of support feature mining block}
	 \setlength{\tabcolsep}{6mm}
	  \begin{tabular}{cccccc}
	  \toprule
	  \multirow{1.5}[2]{*}{\textbf{Method}} & \multirow{1.5}[2]{*}{\textbf{$\bm{AP^{box}}$}} & \multirow{1.5}[2]{*}{\textbf{$\bm{AP^{box}_{50}}$}} & \multirow{1.5}[2]{*}{\textbf{$\bm{AP^{box}_{75}}$}} & \multirow{1.5}[2]{*}{\textbf{FPS}} & \multirow{1.5}[2]{*}{\makecell[c]{\textbf{Model Size}\\\textbf{(MB)}}} \\
			&       &       &       &       &  \\
	  \midrule
	  CBAM \cite{CBAM} & 49.8 & 75.1 & 59.7 & 43   &18.6  \\[2 ex]
	  CoTNet \cite{ContextualTN}   & 51.7 & 76.7 & 62.2 & 48   & 21.0 \\[2 ex]
	  Polarized Self-Attention \cite{PolarizedST} & 51.5 & 76.6 & 61.3 & 45    & 19.8 \\[2 ex]
	  \textbf{OreFSDet(ours)}  &\textbf{52.5}  & \textbf{77.9 }&\textbf{63.0}  & \textbf{50}    & \textbf{19.4} \\
	  \bottomrule
	  \end{tabular}%
	\label{sm}%
  \end{table}%
\subsubsection{Impact of SM}
\par We employ three distinct attention mechanisms to efficiently characterize the feature information of support features as given in Table \ref{sm}.
Convolutional attention module (CBAM \cite{CBAM}) that integrates channel and spatial attention processes. 
Based on the dual attention process, Polarized self-attention \cite{PolarizedST} suggests a more refined version of polarized self-attention. 
CoTNet \cite{ContextualTN} is a transformer-style module that fully exploits the context information contained between input keys to direct the learning of a dynamic attention matrix.
In contrast to the approaches mentioned above, our approach utilizes linear projection to encode feature expression along the two dimensions of height and width rather than 2D convolution or an attention mechanism. 
The results demonstrate that SM Block outperforms other approaches due to its quicker inference speed and better precision.


\begin{table}[!t]
	\renewcommand{\arraystretch}{1.3}
	\centering
	\caption{The effectiveness of relationship guidance block}
	 \setlength{\tabcolsep}{6mm}
	  \begin{tabular}{ccccccc}
	  \toprule
	  \multirow{1.5}[2]{*}{\textbf{Method}} & \multirow{1.5}[2]{*}{\textbf{$\bm{AP^{box}}$}} & \multirow{1.5}[2]{*}{\textbf{$\bm{AP^{box}_{50}}$}} & \multirow{1.5}[2]{*}{\textbf{$\bm{AP^{box}_{75}}$}} & \multirow{1.5}[2]{*}{\textbf{$\bm{AP^{m}_{ }}$}} & \multirow{1.5}[2]{*}{\textbf{$\bm{AP^{l}_{ }}$}} & \multirow{1.5}[2]{*}{\textbf{FPS}} \\
			&       &       &       &       &       &  \\
	  \midrule
	  AttentionRPN \cite{AttentionRPN} & 50.8 & 76.3 & 60.4 & 51.1 & 55.5 & 37 \\[2 ex]
	  FGN \cite{FGN}   & 51.7 & 77.4 & 61.8  & 52.0 & 55.4  & \textbf{50}  \\[2 ex]
	  DAnA \cite{Dual_Awareness}  & 45.2  & 72.9 & 51.5 & 45.4  & 53.9 & 40 \\[2 ex]
	  \textbf{OreFSDet(ours)}   &\textbf{52.5}  & \textbf{77.9} & \textbf{63.0} &\textbf{52.7}  & \textbf{56.0} &\textbf{50}  \\
	  \bottomrule
	  \end{tabular}%
	\label{rg}%
\end{table}%

\subsubsection{Impact of RG}
\par We use four different few-shot strategies to establish the relationship between support and query features. 
The support feature is encoded into a class attention vector in \cite{FGN}, and then element-by-element multiplication is performed along the channel dimension at each spatial position of the query feature map. 
AttentionRPN \cite{AttentionRPN} takes support feature map as the convolution kernel, 
and performs sliding convolution on the query feature to establish the association between the support feature and the query feature. 
DAnA \cite{Dual_Awareness} converts the support feature into a query-position-aware (QPA) feature with specific semantic information, 
and areas of the query feature map with high QPA response should be identified as targets. 
When used for dense ore detection, this inappropriate treatment of spatial characteristics results in poor performance, especially speed burdens. 
Unlike AttentionRPN, our method uses kernels with different sizes of support spatial information, which are fully utilized to obtain spatial correlation. 
And two features are concatenated in the channel dimension to obtain the feature channel correlation. 
The results show that our method outperforms the other three methods and is lightweight enough without additional speed and model size burden.

\subsubsection{Impact of Dual-scale Semantic Aggregation Module}
\begin{figure}[!t]    
	\centerline{\includegraphics[width=3in]{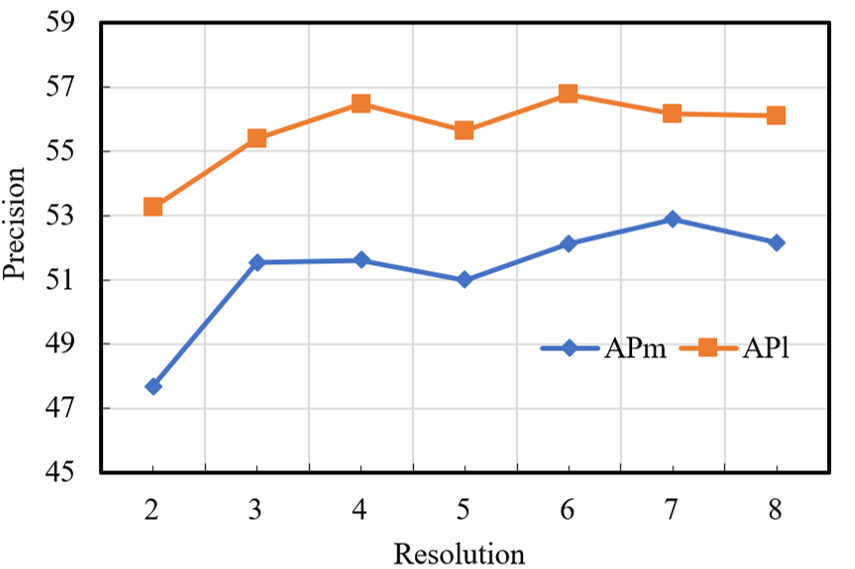}}   
	\caption{ The performance of different resolutions on ore dataset.
	}
	\label{resolution}
\end{figure}

The impact of resolution on model performance can be seen in Fig. \ref{resolution}. 
With the decrease of resolution, the overall $AP^{box}$ of the model does not change much, 
and the corresponding $AP^{m}$ shows a stable downward trend. 
The $AP^{l}$ nevertheless maintains a fairly high level with resolution = 3, which indicates that small resolution does tend to large target information. 
As shown in Table \ref{ROI}, our proposed dual-scale semantic aggregation module takes into account the detection of different scales of ore, 
and has a significant improvement in performance. Interestingly, when aggregating two smaller resolution feature maps, 
we achieved a smaller model size while sacrificing only a tiny amount of speed.

\begin{table}[!t]
	\renewcommand{\arraystretch}{1.3}
	\centering
	\caption{The effectiveness of dual-scale semantic aggregation module}
	 \setlength{\tabcolsep}{5mm}
	  \begin{tabular}{cccccccc}
	  \toprule
	   \multirow{1.5}[2]{*}{\textbf{Resolution}} & \multirow{1.5}[2]{*}{\textbf{$\bm{AP^{box}_{75}}$}} & \multirow{1.5}[2]{*}{\textbf{$\bm{AP^{box}_{50}}$}} & \multirow{1.5}[2]{*}{\textbf{$\bm{AP^{box}_{75}}$}} & \multirow{1.5}[2]{*}{\textbf{$\bm{AP^{m}_{ }}$}} & \multirow{1.5}[2]{*}{\textbf{$\bm{AP^{l}_{ }}$}} & \multirow{1.5}[2]{*}{\textbf{FPS}}& \multirow{1.5}[2]{*}{\makecell[c]{\textbf{Model Size}\\\textbf{(MB)}}}\\
			&       &       &       &       &       &  \\
	  \midrule
	  3     & 51.4 & 77.2 & 61.6  & 51.5 & 55.4  & 48&15.9 \\[2 ex]
	  4     & 51.5 & 77.0 & 61.7 & 51.6 &\textbf{56.5}  & 48& \textbf{16.4}\\[2 ex]
	  5     & 51.6 & 76.7 & 61.7 & 51.9 & 54.1 & 48&16.9\\[2 ex]
	  8     & 52.0 & 76.8 & 62.1 & 52.2 & 56.1  & 48&19.4\\[2 ex]
	 \textbf{4\&8}  & \textbf{52.5} & \textbf{77.9} & \textbf{63.0} & \textbf{52.7} & 56.0 & \textbf{50}& 19.4\\[2 ex]
	 \textbf{3\&5}  & \textbf{52.3} & \textbf{78.1} & \textbf{62.6} & \textbf{52.4} & 56.3 & \textbf{48}& 16.5\\
	  \bottomrule
	  \end{tabular}%
	\label{ROI}%
  \end{table}%

\subsubsection{Visualization}
\begin{figure}[!t]    
	\centerline{\includegraphics[width=6in]{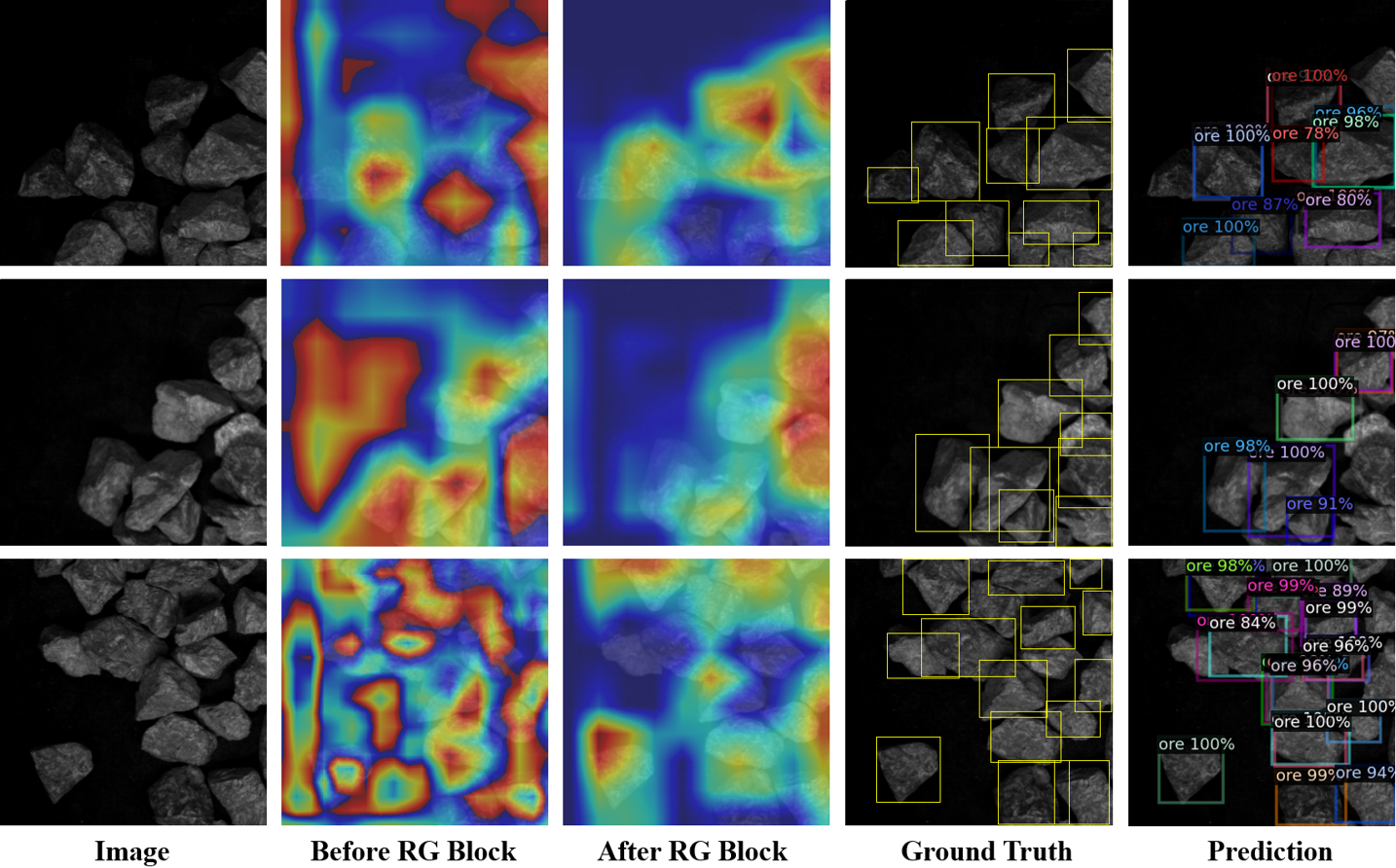}}   
	\caption{ The visualization of before and after RG Block.
	}
	\label{RG_visual} 
\end{figure}

\begin{figure}[!t]    
	\centerline{\includegraphics[width=4in]{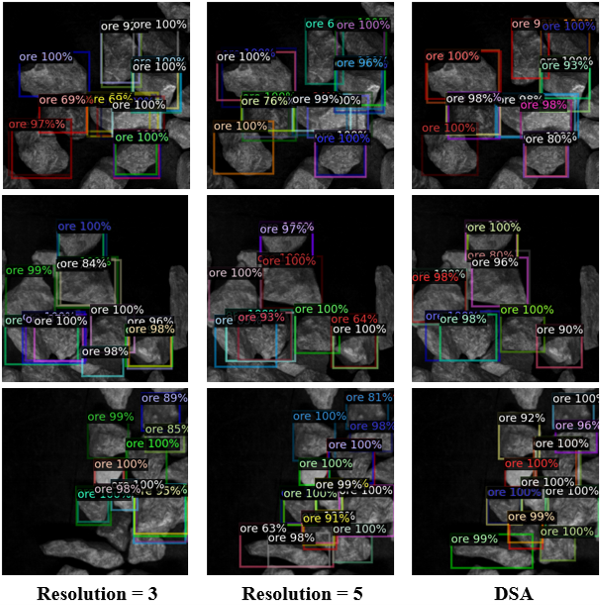}}    
	\caption{ The visualization of the effect of DSA.
	}
	\label{resolution_visual}
\end{figure}

To further illustrate the effect of the RG Block, we visualize the features before and after it. 
As shown in Fig. \ref{RG_visual}, after passing through the module, the query features get activated to focus on more important features, rather than the previously cluttered state.
Our proposed RG Block fully establishes the spatial correlation and channel correlation between support and query features, which significantly improve the guidance performance.
Moreover, the visualization of the effect of the DSA module is presented in Fig. \ref{resolution_visual}.
Ores of different sizes have different sensitivities to resolution, 
which leads to different confidence scores. 
DSA combines two resolutions to take into account the detection of different sizes of ore and achieved better performance. 
Results demonstrate that DSA can effectively retrieve detailed features at different resolutions to contribute with the prediction process.
Finally, we visualize different results of few-shot object detection from ore images. Compared with other detection methods, the detection results obtained by OreFSDet have high confidence and no excessive miss detection.

\subsection{Comparison with State-of-the-Art Methods}
On the ore dataset, we compare the performance of several FSOD algorithms in terms of $AP^{box}$, 
${AP^{box}_{50}}$ and ${AP^{box}_{75}}$. 
TFA \cite{TFA} trains the detector on MS COCO and then fine-tunes the last layer of the detector on a
small balanced ore dataset while freezing other parameters of the model.
FSCE \cite{FSCE} provides a comparative learning method into the two-stage fine-tuning method to reduce the intra-class
differences and increase the inter-class difference.
Support features and query features are fused after the RPN for guiding detection
in two-stage methods MetaRCNN \cite{Meta_RCNN} and FsDetView \cite{FSDetView}, while
AttentionRPN [7] provided a feature fusion module to exclude proposals by class category before RPN.
MPSR \cite{MPSR} presented a multi-scale positive sample refining technique to enrich the object scale.
Our proposed OreFSDet is also a fine-tuning method. 
Different from the above methods, it mines the important features of ore images through SM Block, 
thereby removing the detection interference caused by the introduction of background noise. 
Then, the spatial and channel correlation for precise guidance between the support image and query image are fully established through the attention mechanism. 
Finally, the dual-scale semantic aggregation module retrieves detailed features at different resolutions to contribute with the prediction process.
\begin{table}[!t]
	\renewcommand{\arraystretch}{1.6}
	\caption{Comparative experiments for different FSOD algorithms on the ore dataset under different shots}
	\label{7}
	\centering
	\small
	\setlength{\tabcolsep}{2.5mm}{
		\begin{tabular}{c|ccc|ccc|ccc} 
      \toprule
      \multirow{2}{*}{\textbf{Method}} & \multicolumn{3}{c|}{\textbf{5-shot}} & \multicolumn{3}{c|}{\textbf{15-shot}} & \multicolumn{3}{c}{\textbf{25-shot}} \\
      \cmidrule(r){2-4} \cmidrule(r){4-6} \cmidrule(r){6-10}
      & $\bm{AP^{box}}$ & $AP_{50}^{box}$ & $AP_{75}^{box}$ & $\bm{AP^{box}}$ & $AP_{50}^{box}$ & $AP_{75}^{box}$ & $\bm{AP^{box}}$ & $AP_{50}^{box} $ & $AP_{75}^{box} $ \\
      \midrule
      TFA \cite{TFA}   &  28.9     &   48  & 33.6  &    30.5  &  49     & 35.6&   31.5    &    49.7   & 36.5 \\
      FSCE \cite{FSCE} &    33.5   &    47.2   & 40.8  &    35.3 &  49   & 42.8 &    37.0  &   50.4    & 44.5 \\
      Meta R-CNN \cite{Meta_RCNN} &   15.4    &    34.7   & 9.4  &  17.6     &   37.3    & 13.0  &    22.0   &  39.0    & 24.1 \\
      FSDetView \cite{FSDetView} &    18.2   &   36.1    & 15.4   &    20.1   &   38.6    & 17.7 & 25.4     &   41.1    & 29.8 \\
      MPSR \cite{MPSR}     &   32.4    & 45.8  &   39.7    &   34.2   & 47.6  &  41.8     &   37.0    & 52.1& 44.3\\
      AttentionRPN \cite{AttentionRPN} &   25.1    &    44.0   & 27.0  &   29.2    &  45.9     & 34.5  &  30.8    &47.3   & 37.0 \\
      OreFSDet(ours)  &   \textbf{48.5}   &  \textbf{74.1}     & \textbf{57.6}  &  \textbf{52.1}     &   \textbf{77.2}     & \textbf{62.5}  &  \textbf{54.1}    &    \textbf{78.4}  & \textbf{64.7} \\
      \bottomrule
      \end{tabular}}
\end{table}

As shown in Table \ref{7}, we conducted Comparative experiments for different FSOD algorithms on the ore dataset under different shots. With 25 shots provided for training and 1060 ore images for assessment,
our method surpasses the baseline \cite{AttentionRPN} with great advantage by 23.3/31.1/27.7 respectively on $AP^{box}$/${AP^{box}_{50}}$/${AP^{box}_{75}}$ metrics. 
Additionally, with fewer shots, the suggested model remains excellent performance, 
proving that OreFSDet effectively retrieves the information of support images  for guidance through SG Block, RG Block, and the dual-scale semantic aggregation module.

\begin{table}[!t]
	\renewcommand{\arraystretch}{1.3}
	\caption{Comparison with the state-of-the-art methods on ore dataset
		\label{8}}
	\centering
	\small 
	
	\setlength{\tabcolsep}{2.2mm}{
		\begin{tabular}{lcccccccc}
			\toprule
			\multirow{2}{*}{\textbf{Method}}& 
			\multirow{2}{*}{\textbf{Backbone}}& 
			\multirow{2}{*}{\textbf{$\bm{AP^{box}}$}}	& 
      \multirow{2}{*}{\textbf{$\bm{AP_{50}^{box}} $}}	& 
			\multirow{2}{*}{\textbf{$\bm{AP_{75}^{box}} $}}	& 
			\multirow{2}{*}{\textbf{FPS}} & 
			\multirow{2}{*}{\makecell[c]{\textbf{Model Size}\\\textbf{(MB)}}}&\multirow{2}{*}{\makecell[c]{\textbf{Inf.Memory}\\\textbf{(MB)}}} 
			\\ 
			& & & & & & &  \\   
			\midrule 
			\emph{\textbf{General object detection}}  & & & & & &  \\
			Faster R-CNN \cite{Faster_RCNN}	&ResNet50~\cite{Deep_Residual} &42.6 &51.7 &48.6 &11 &264 &5221 \\
            Faster R-CNN \cite{Faster_RCNN}	&ResNet18~\cite{Deep_Residual} &42.6 &51.6 &48.6 &43 &93 &1277 \\
            Faster R-CNN \cite{Faster_RCNN}	&V-19-Slim\cite{vovnet} &40.6&51.3 &47.2 &50 &\textbf{51} &1269 \\
            SSD \cite{SSD}	&VGG16\cite{VGG} &41.7  &50.8  &48.3 &\textbf{77} &190 &9787 \\
			RetinaNet \cite{RetinaNet} & ResNet50~\cite{Deep_Residual}  &41.1 &57.7 &46.9 &26 &290 &1337 \\ 
            RetinaNet \cite{RetinaNet} & SwinT~\cite{swin} & 42.1  & 56.5  & 47.7  & 41 & 282   & 1757 \\
            RetinaNet \cite{RetinaNet} & PVTv2~\cite{PVTv2} & 41.6  & 55.4  & 47.4  & 51 & 130   & 1587 \\
            RetinaNet \cite{RetinaNet} & ResNet18~\cite{Deep_Residual} & 41.7  & 57.3  & 47.3  & 50 & 91   & 1543 \\
			Cascade R-CNN \cite{Cascade_R_CNN}        &ResNet50~\cite{Deep_Residual} &43.6&51.0 &49.6 &21 &553 &1699  \\
            YOLOv3 \cite{yolov3}        &Darknet\cite{yolov3}   & 37.4  & 50.7&47.3 &48 &493 & 1555\\
			Grid R-CNN \cite{Grid_R_CNN}	&ResNet50~\cite{Deep_Residual}  &43.8 &54.0 &\textbf{50.0} &22 &515 &1713 \\
            CenterNet2 \cite{centernet2}        &DLA~\cite{dla}  &43.6 &55.6 &47.4 &38 &357 &1339 \\
			FCOS \cite{FCOS}		        &ResNet50~\cite{Deep_Residual}  &42.8 &55.4 &48.4 &29 &256 &\textbf{1303} \\
			VarifocalNet \cite{VarifocalNet}        &ResNet50~\cite{Deep_Residual}  &45.0 &63.6 &48.6 &24 &261 &1479 \\
			YOLOF \cite{YOLOf}        &ResNet50~\cite{Deep_Residual}  &\textbf{46.8}   &\textbf{67.5} &49.3 &45 &338 &1433 \\
            DDOD \cite{ddod}  & ResNet50~\cite{Deep_Residual} & 44.0    & 53.9  & 49.4  & 50 & 245   & 1719 \\
            GFL \cite{gfl}  & ResNet50~\cite{Deep_Residual} & 45.8  & 63.0    & 49.6  & 50 & 246   & 1709 \\
            PAA \cite{paa}  & ResNet50~\cite{Deep_Residual} & 46.0    & 65.2  & 49.1  & 22 & 245   & 1661 \\

      \midrule
			\emph{\textbf{Few-shot object detection$^*$}} & & & & & &  \\
			TFA \cite{TFA}		 &ResNet101~\cite{Deep_Residual} &31.5 &49.7 &36.5 &16 &230 &2013 \\
			Meta R-CNN	\cite{Meta_RCNN}	&ResNet101~\cite{Deep_Residual} &22.0 &39.0 &24.1 &28 &148 &2639 \\
			FSDetView \cite{FSDetView}	&ResNet101~\cite{Deep_Residual} &25.4 &41.1 &29.8 &29 &157 &2665 \\
			AttentionRPN 	\cite{AttentionRPN}	        &ResNet50~\cite{Deep_Residual} &30.8 &47.3 &37.0 &28&211 &1667 \\
            MPSR 	  \cite{MPSR}       	&ResNet101~\cite{Deep_Residual} &37.1 &52.1 &44.3 &21 &462 &3821  \\
            MPSR 	  \cite{MPSR}  &V-19-Slim\cite{vovnet} &24.4 &46.2 &23.9 &36 & 117& 1941\\
            FSCE  \cite{FSCE}		&ResNet101~\cite{Deep_Residual} &37.0 &50.4 &44.5 &18 &298 &2200 \\
            FSCE  \cite{FSCE}		&ResNet18~\cite{Deep_Residual} &23.9 &54.3 &18.4 &\textbf{55} &51 &1637 \\
            FSCE \cite{FSCE}  &V-19-Slim\cite{vovnet} &28.9 &55.2 &29.1 &43 & 51& 1611\\
            
            OreFSDet(ours)  &V-19-Slim\cite{vovnet} &\textbf{54.1} &\textbf{78.4} &\textbf{64.7} &50 & \textbf{19}& \textbf{1059}\\
			\bottomrule
	\end{tabular}}   
        	\begin{tablenotes}
			\footnotesize
			\item[]$*$: The methods are implemented under the 25-shot setting.
		\end{tablenotes}
\end{table}

\begin{figure}[!t]    
	\centerline{\includegraphics[width=6in]{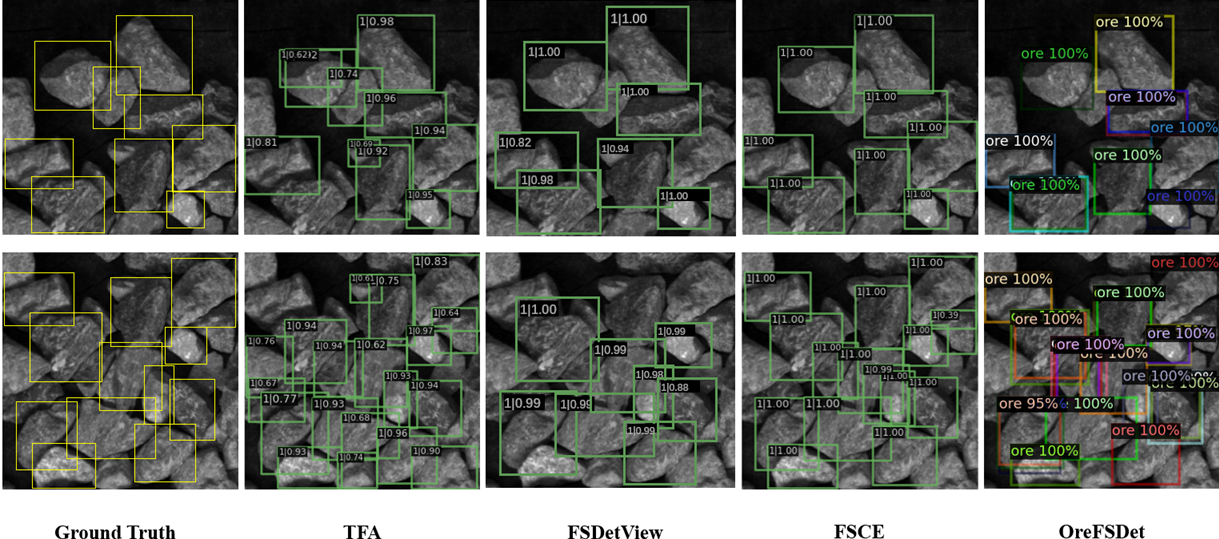}}   
	\caption{ The different visual results of few-shot object detection from ore images.
	}
	\label{visual_result} 
\end{figure}
The comparison of each approach under FSOD and general object detection is shown in Table \ref{8}.
General object detection methods with sufficient training data tend to achieve better performance than 
few-shot object detection methods by learning a large amount of information between categories.
In particular, many transformer-based methods have recently been proposed to solve various computer vision tasks such as Swin transformers \cite{swin}.
To obtain a lightweight model and faster detection speed, we develop a few-shot object detector on CenterNet2 \cite{centernet2} instead of traditional Faster RCNN \cite{Faster_RCNN}.
In Table \ref{8}, OreFSDet not only rivals the best detector of general object detection (even better than YOLOF by 5.7/10.4 on $AP^{box}$/${AP^{box}_{75}}$ metrics), 
but also consistently beats the few-shot detectors with a great performance gap on all metrics. 
In addition, OreFSDet performs best at a model size of 19MB as well as being competitive at 50 FPS among general object detectors.


%% file: files/5-Conclusion.tex
\section{Conclusion}

In this work, we observe that the FSOD task for ore images is extremely difficult to be utilized in a real-world scenario deployment due to high computational overhead and memory requirements.
To this end, we present a lightweight and effective framework OreFSDet to solve the problems. 
On the ore dataset, our model exceeds the best in the FSOD algorithms with great advantage by 25.5/21 on $AP^{box}$/${FPS}$ metrics, respectively. 
Moreover, our OreFSDet performs best with the model size of 19MB and is competitive at 50 FPS among general object detectors.
Our method is advantageous in a lightweight model and ultra-fast detection speed and can be effectively deployed in mineral processing operations.
Furthermore, with only a few samples, our method can be extended to some objects that share similar properties to the ores. 

For future work, we will focus on extending our method to the more challenging few-shot instance segmentation and one-shot object detection by new mechanisms and networks.
\\

%% file: cas-sc-template.bbl
\begin{thebibliography}{45}
\expandafter\ifx\csname natexlab\endcsname\relax\def\natexlab#1{#1}\fi
\providecommand{\url}[1]{\texttt{#1}}
\providecommand{\href}[2]{#2}
\providecommand{\path}[1]{#1}
\providecommand{\DOIprefix}{doi:}
\providecommand{\ArXivprefix}{arXiv:}
\providecommand{\URLprefix}{URL: }
\providecommand{\Pubmedprefix}{pmid:}
\providecommand{\doi}[1]{\href{http://dx.doi.org/#1}{\path{#1}}}
\providecommand{\Pubmed}[1]{\href{pmid:#1}{\path{#1}}}
\providecommand{\bibinfo}[2]{#2}
\ifx\xfnm\relax \def\xfnm[#1]{\unskip,\space#1}\fi
\bibitem[{Cai and Vasconcelos(2021)}]{Cascade_R_CNN}
\bibinfo{author}{Cai, Z.}, \bibinfo{author}{Vasconcelos, N.},
  \bibinfo{year}{2021}.
\newblock \bibinfo{title}{Cascade r-cnn: High quality object detection and
  instance segmentation}.
\newblock \bibinfo{journal}{IEEE Transactions on Pattern Analysis and Machine
  Intelligence} \bibinfo{volume}{43}, \bibinfo{pages}{1483--1498}.
\bibitem[{Chen et~al.(2021a)Chen, Wang, Yang, Zhang, Cheng and Sun}]{YOLOf}
\bibinfo{author}{Chen, Q.}, \bibinfo{author}{Wang, Y.}, \bibinfo{author}{Yang,
  T.}, \bibinfo{author}{Zhang, X.}, \bibinfo{author}{Cheng, J.},
  \bibinfo{author}{Sun, J.}, \bibinfo{year}{2021}a.
\newblock \bibinfo{title}{You only look one-level feature}, in:
  \bibinfo{booktitle}{IEEE/CVF Conference on Computer Vision and Pattern
  Recognition}, pp. \bibinfo{pages}{13034 -- 13043}.
\bibitem[{Chen et~al.(2021b)Chen, Liu, Su, Chang, Lin, Yeh, Chen and
  Hsu}]{Dual_Awareness}
\bibinfo{author}{Chen, T.I.}, \bibinfo{author}{Liu, Y.C.}, \bibinfo{author}{Su,
  H.T.}, \bibinfo{author}{Chang, Y.C.}, \bibinfo{author}{Lin, Y.H.},
  \bibinfo{author}{Yeh, J.F.}, \bibinfo{author}{Chen, W.C.},
  \bibinfo{author}{Hsu, W.}, \bibinfo{year}{2021}b.
\newblock \bibinfo{title}{Dual-awareness attention for few-shot object
  detection}.
\newblock \bibinfo{journal}{IEEE Transactions on Multimedia} ,
  \bibinfo{pages}{1--11}.
\bibitem[{Chen et~al.(2021c)Chen, Yang, Li, Zhao, Zha and Wu}]{ddod}
\bibinfo{author}{Chen, Z.}, \bibinfo{author}{Yang, C.}, \bibinfo{author}{Li,
  Q.}, \bibinfo{author}{Zhao, F.}, \bibinfo{author}{Zha, Z.J.},
  \bibinfo{author}{Wu, F.}, \bibinfo{year}{2021}c.
\newblock \bibinfo{title}{Disentangle your dense object detector}, in:
  \bibinfo{booktitle}{Proceedings of the 29th ACM International Conference on
  Multimedia}, pp. \bibinfo{pages}{4939--4948}.
\bibitem[{Cui et~al.(2022)Cui, Liao, Hu, An and Liu}]{CUI2022108296}
\bibinfo{author}{Cui, Y.}, \bibinfo{author}{Liao, Q.}, \bibinfo{author}{Hu,
  D.}, \bibinfo{author}{An, W.}, \bibinfo{author}{Liu, L.},
  \bibinfo{year}{2022}.
\newblock \bibinfo{title}{Coarse-to-fine pseudo supervision guided meta-task
  optimization for few-shot object classification}.
\newblock \bibinfo{journal}{Pattern Recognition} \bibinfo{volume}{122},
  \bibinfo{pages}{108296}.
\bibitem[{Fan et~al.(2020a)Fan, Zhuo, Tang and Tai}]{AttentionRPN}
\bibinfo{author}{Fan, Q.}, \bibinfo{author}{Zhuo, W.}, \bibinfo{author}{Tang,
  C.K.}, \bibinfo{author}{Tai, Y.W.}, \bibinfo{year}{2020}a.
\newblock \bibinfo{title}{Few-shot object detection with attention-rpn and
  multi-relation detector}, in: \bibinfo{booktitle}{IEEE/CVF Conference on
  Computer Vision and Pattern Recognition}, pp. \bibinfo{pages}{4012 --4021}.
\bibitem[{Fan et~al.(2020b)Fan, Yu, Liang, Ou, Gao, Xia and Li}]{FGN}
\bibinfo{author}{Fan, Z.}, \bibinfo{author}{Yu, J.G.}, \bibinfo{author}{Liang,
  Z.}, \bibinfo{author}{Ou, J.}, \bibinfo{author}{Gao, C.},
  \bibinfo{author}{Xia, G.S.}, \bibinfo{author}{Li, Y.}, \bibinfo{year}{2020}b.
\newblock \bibinfo{title}{Fgn: Fully guided network for few-shot instance
  segmentation}, in: \bibinfo{booktitle}{IEEE/CVF Conference on Computer Vision
  and Pattern Recognition}, pp. \bibinfo{pages}{9169--9178}.
\bibitem[{He et~al.(2016)He, Zhang, Ren and Sun}]{Deep_Residual}
\bibinfo{author}{He, K.}, \bibinfo{author}{Zhang, X.}, \bibinfo{author}{Ren,
  S.}, \bibinfo{author}{Sun, J.}, \bibinfo{year}{2016}.
\newblock \bibinfo{title}{Deep residual learning for image recognition}, in:
  \bibinfo{booktitle}{IEEE Conference on Computer Vision and Pattern
  Recognition}, pp. \bibinfo{pages}{770--778}.
\bibitem[{Kang et~al.(2019)Kang, Liu, Wang, Yu, Feng and Darrell}]{METAYOLO}
\bibinfo{author}{Kang, B.}, \bibinfo{author}{Liu, Z.}, \bibinfo{author}{Wang,
  X.}, \bibinfo{author}{Yu, F.}, \bibinfo{author}{Feng, J.},
  \bibinfo{author}{Darrell, T.}, \bibinfo{year}{2019}.
\newblock \bibinfo{title}{Few-shot object detection via feature reweighting}.
\newblock \bibinfo{journal}{IEEE/CVF International Conference on Computer
  Vision} , \bibinfo{pages}{8419--8428}.
\bibitem[{Kim et~al.(2021)Kim, Jung and Lee}]{Spatial}
\bibinfo{author}{Kim, G.}, \bibinfo{author}{Jung, H.G.}, \bibinfo{author}{Lee,
  S.W.}, \bibinfo{year}{2021}.
\newblock \bibinfo{title}{Spatial reasoning for few-shot object detection}.
\newblock \bibinfo{journal}{Pattern Recognition} \bibinfo{volume}{120},
  \bibinfo{pages}{108118}.
\bibitem[{Kim and Lee(2020)}]{paa}
\bibinfo{author}{Kim, K.J.}, \bibinfo{author}{Lee, H.S.}, \bibinfo{year}{2020}.
\newblock \bibinfo{title}{Probabilistic anchor assignment with iou prediction
  for object detection}, in: \bibinfo{booktitle}{European Conference on
  Computer Vision}, pp. \bibinfo{pages}{355--371}.
\bibitem[{Lee et~al.(2019)Lee, Hwang, Lee, Bae and Park}]{vovnet}
\bibinfo{author}{Lee, Y.}, \bibinfo{author}{Hwang, J.}, \bibinfo{author}{Lee,
  S.}, \bibinfo{author}{Bae, Y.}, \bibinfo{author}{Park, J.},
  \bibinfo{year}{2019}.
\newblock \bibinfo{title}{An energy and gpu-computation efficient backbone
  network for real-time object detection}, pp. \bibinfo{pages}{752--760}.
\bibitem[{Li et~al.(2020a)Li, Pan, Chen, Wulamu and Yang}]{UNet}
\bibinfo{author}{Li, H.}, \bibinfo{author}{Pan, C.}, \bibinfo{author}{Chen,
  Z.}, \bibinfo{author}{Wulamu, A.}, \bibinfo{author}{Yang, A.},
  \bibinfo{year}{2020}a.
\newblock \bibinfo{title}{Ore image segmentation method based on u-net and
  watershed}.
\newblock \bibinfo{journal}{Computers, Materials and Continua}
  \bibinfo{volume}{65}, \bibinfo{pages}{563--578}.
\bibitem[{Li et~al.(2020b)Li, Wang, Wu, Chen, Hu, Li, Tang and Yang}]{gfl}
\bibinfo{author}{Li, X.}, \bibinfo{author}{Wang, W.}, \bibinfo{author}{Wu, L.},
  \bibinfo{author}{Chen, S.}, \bibinfo{author}{Hu, X.}, \bibinfo{author}{Li,
  J.}, \bibinfo{author}{Tang, J.}, \bibinfo{author}{Yang, J.},
  \bibinfo{year}{2020}b.
\newblock \bibinfo{title}{Generalized focal loss: Learning qualified and
  distributed bounding boxes for dense object detection}, in:
  \bibinfo{booktitle}{Advances in Neural Information Processing Systems}, pp.
  \bibinfo{pages}{1--11}.
\bibitem[{Li et~al.(2023a)Li, Yao, Pan and Mei}]{ContextualTN}
\bibinfo{author}{Li, Y.}, \bibinfo{author}{Yao, T.}, \bibinfo{author}{Pan, Y.},
  \bibinfo{author}{Mei, T.}, \bibinfo{year}{2023}a.
\newblock \bibinfo{title}{Contextual transformer networks for visual
  recognition}.
\newblock \bibinfo{journal}{IEEE Transactions on Pattern Analysis and Machine
  Intelligence} \bibinfo{volume}{45}, \bibinfo{pages}{1489--1500}.
\bibitem[{Li et~al.(2023b)Li, Hu, Luo and Hu}]{SaberNet}
\bibinfo{author}{Li, Z.}, \bibinfo{author}{Hu, Z.}, \bibinfo{author}{Luo, W.},
  \bibinfo{author}{Hu, X.}, \bibinfo{year}{2023}b.
\newblock \bibinfo{title}{Sabernet: Self-attention based effective relation
  network for few-shot learning}.
\newblock \bibinfo{journal}{Pattern Recognition} \bibinfo{volume}{133},
  \bibinfo{pages}{109024}.
\bibitem[{Lin et~al.(2020)Lin, Goyal, Girshick, He and Dollár}]{RetinaNet}
\bibinfo{author}{Lin, T.Y.}, \bibinfo{author}{Goyal, P.},
  \bibinfo{author}{Girshick, R.}, \bibinfo{author}{He, K.},
  \bibinfo{author}{Dollár, P.}, \bibinfo{year}{2020}.
\newblock \bibinfo{title}{Focal loss for dense object detection}.
\newblock \bibinfo{journal}{IEEE Transactions on Pattern Analysis and Machine
  Intelligence} \bibinfo{volume}{42}, \bibinfo{pages}{318--327}.
\bibitem[{Lin et~al.(2014)Lin, Maire, Belongie, Hays, Perona, Ramanan,
  Doll{\'a}r and Zitnick}]{coco}
\bibinfo{author}{Lin, T.Y.}, \bibinfo{author}{Maire, M.},
  \bibinfo{author}{Belongie, S.}, \bibinfo{author}{Hays, J.},
  \bibinfo{author}{Perona, P.}, \bibinfo{author}{Ramanan, D.},
  \bibinfo{author}{Doll{\'a}r, P.}, \bibinfo{author}{Zitnick, C.L.},
  \bibinfo{year}{2014}.
\newblock \bibinfo{title}{Microsoft coco: Common objects in context}, in:
  \bibinfo{booktitle}{European Conference on Computer Vision}, pp.
  \bibinfo{pages}{740--755}.
\bibitem[{Liu et~al.(2022)Liu, Liu, Fan and Huang}]{PolarizedST}
\bibinfo{author}{Liu, H.}, \bibinfo{author}{Liu, F.}, \bibinfo{author}{Fan,
  X.}, \bibinfo{author}{Huang, D.}, \bibinfo{year}{2022}.
\newblock \bibinfo{title}{Polarized self-attention: Towards high-quality
  pixel-wise regression}.
\newblock \bibinfo{journal}{Neurocomputing} \bibinfo{volume}{506},
  \bibinfo{pages}{158--167}.
\bibitem[{Liu et~al.(2016)Liu, Anguelov, Erhan, Szegedy, Reed, Fu and
  Berg}]{SSD}
\bibinfo{author}{Liu, W.}, \bibinfo{author}{Anguelov, D.},
  \bibinfo{author}{Erhan, D.}, \bibinfo{author}{Szegedy, C.},
  \bibinfo{author}{Reed, S.}, \bibinfo{author}{Fu, C.Y.},
  \bibinfo{author}{Berg, A.C.}, \bibinfo{year}{2016}.
\newblock \bibinfo{title}{Ssd: Single shot multibox detector}, in:
  \bibinfo{booktitle}{European Conference on Computer Vision}, pp.
  \bibinfo{pages}{21--37}.
\bibitem[{Liu et~al.(2020)Liu, Zhang, Jing, Wang and Zhao}]{LiuXiaobo2020OreIS}
\bibinfo{author}{Liu, X.}, \bibinfo{author}{Zhang, Y.}, \bibinfo{author}{Jing,
  H.}, \bibinfo{author}{Wang, L.}, \bibinfo{author}{Zhao, S.},
  \bibinfo{year}{2020}.
\newblock \bibinfo{title}{Ore image segmentation method using u-net and
  res\_unet convolutional networks}.
\newblock \bibinfo{journal}{RSC Advances} \bibinfo{volume}{10},
  \bibinfo{pages}{9396--9406}.
\bibitem[{Liu et~al.(2021a)Liu, Zhang, Liu, Wang and Xia}]{oresegmentation}
\bibinfo{author}{Liu, Y.}, \bibinfo{author}{Zhang, Z.}, \bibinfo{author}{Liu,
  X.}, \bibinfo{author}{Wang, L.}, \bibinfo{author}{Xia, X.},
  \bibinfo{year}{2021}a.
\newblock \bibinfo{title}{Efficient image segmentation based on deep learning
  for mineral image classification}.
\newblock \bibinfo{journal}{Advanced Powder Technology} \bibinfo{volume}{32},
  \bibinfo{pages}{3885--3903}.
\bibitem[{Liu et~al.(2021b)Liu, Lin, Cao, Hu, Wei, Zhang, Lin and Guo}]{swin}
\bibinfo{author}{Liu, Z.}, \bibinfo{author}{Lin, Y.}, \bibinfo{author}{Cao,
  Y.}, \bibinfo{author}{Hu, H.}, \bibinfo{author}{Wei, Y.},
  \bibinfo{author}{Zhang, Z.}, \bibinfo{author}{Lin, S.}, \bibinfo{author}{Guo,
  B.}, \bibinfo{year}{2021}b.
\newblock \bibinfo{title}{Swin transformer: Hierarchical vision transformer
  using shifted windows}, in: \bibinfo{booktitle}{IEEE/CVF International
  Conference on Computer Vision}, pp. \bibinfo{pages}{9992--10002}.
\bibitem[{Lu et~al.(2019)Lu, Li, Yue, Li and Yan}]{Grid_R_CNN}
\bibinfo{author}{Lu, X.}, \bibinfo{author}{Li, B.}, \bibinfo{author}{Yue, Y.},
  \bibinfo{author}{Li, Q.}, \bibinfo{author}{Yan, J.}, \bibinfo{year}{2019}.
\newblock \bibinfo{title}{Grid r-cnn}, in: \bibinfo{booktitle}{IEEE/CVF
  Conference on Computer Vision and Pattern Recognition}, pp.
  \bibinfo{pages}{7355--7364}.
\bibitem[{Mukherjee et~al.(2009)Mukherjee, Potapovich, Levner and
  Zhang}]{ore_image}
\bibinfo{author}{Mukherjee, D.P.}, \bibinfo{author}{Potapovich, Y.},
  \bibinfo{author}{Levner, I.}, \bibinfo{author}{Zhang, H.},
  \bibinfo{year}{2009}.
\newblock \bibinfo{title}{Ore image segmentation by learning image and shape
  features}.
\newblock \bibinfo{journal}{Pattern Recognition Letters} \bibinfo{volume}{30},
  \bibinfo{pages}{615--622}.
\bibitem[{Olivier et~al.(2020)Olivier, Maritz and
  Craig}]{Olivier2020EstimatingOP}
\bibinfo{author}{Olivier, L.E.}, \bibinfo{author}{Maritz, M.G.},
  \bibinfo{author}{Craig, I.K.}, \bibinfo{year}{2020}.
\newblock \bibinfo{title}{Estimating ore particle size distribution using a
  deep convolutional neural network}.
\newblock \bibinfo{journal}{IFAC-PapersOnLine} \bibinfo{volume}{53},
  \bibinfo{pages}{12038--12043}.
\bibitem[{Redmon and Farhadi(2018)}]{yolov3}
\bibinfo{author}{Redmon, J.}, \bibinfo{author}{Farhadi, A.},
  \bibinfo{year}{2018}.
\newblock \bibinfo{title}{Yolov3: An incremental improvement}.
\newblock \bibinfo{journal}{arXiv preprint arXiv:1804.02767} .
\bibitem[{Ren et~al.(2017)Ren, He, Girshick and Sun}]{Faster_RCNN}
\bibinfo{author}{Ren, S.}, \bibinfo{author}{He, K.}, \bibinfo{author}{Girshick,
  R.}, \bibinfo{author}{Sun, J.}, \bibinfo{year}{2017}.
\newblock \bibinfo{title}{Faster r-cnn: Towards real-time object detection with
  region proposal networks}.
\newblock \bibinfo{journal}{IEEE Transactions on Pattern Analysis and Machine
  Intelligence} \bibinfo{volume}{39}, \bibinfo{pages}{1137--1149}.
\bibitem[{Shuang et~al.(2021)Shuang, Lyu, Loo and Zhang}]{Scalebalanced}
\bibinfo{author}{Shuang, K.}, \bibinfo{author}{Lyu, Z.}, \bibinfo{author}{Loo,
  J.}, \bibinfo{author}{Zhang, W.}, \bibinfo{year}{2021}.
\newblock \bibinfo{title}{Scale-balanced loss for object detection}.
\newblock \bibinfo{journal}{Pattern Recognition} \bibinfo{volume}{117},
  \bibinfo{pages}{107997}.
\bibitem[{Simonyan and Zisserman(2015)}]{VGG}
\bibinfo{author}{Simonyan, K.}, \bibinfo{author}{Zisserman, A.},
  \bibinfo{year}{2015}.
\newblock \bibinfo{title}{Very deep convolutional networks for large-scale
  image recognition}, in: \bibinfo{booktitle}{International Conference on
  Learning Representations}, pp. \bibinfo{pages}{1--14}.
\bibitem[{Sun et~al.(2021)Sun, Li, Cai, Yuan and Zhang}]{FSCE}
\bibinfo{author}{Sun, B.}, \bibinfo{author}{Li, B.}, \bibinfo{author}{Cai, S.},
  \bibinfo{author}{Yuan, Y.}, \bibinfo{author}{Zhang, C.},
  \bibinfo{year}{2021}.
\newblock \bibinfo{title}{Fsce: Few-shot object detection via contrastive
  proposal encoding}, in: \bibinfo{booktitle}{IEEE/CVF Conference on Computer
  Vision and Pattern Recognition}, pp. \bibinfo{pages}{7348--7358}.
\bibitem[{Sun et~al.(2022)Sun, Huang, Cheng, Jia, Xiong and Zhang}]{Efficient}
\bibinfo{author}{Sun, G.}, \bibinfo{author}{Huang, D.}, \bibinfo{author}{Cheng,
  L.}, \bibinfo{author}{Jia, J.}, \bibinfo{author}{Xiong, C.},
  \bibinfo{author}{Zhang, Y.}, \bibinfo{year}{2022}.
\newblock \bibinfo{title}{Efficient and lightweight framework for real-time ore
  image segmentation based on deep learning}.
\newblock \bibinfo{journal}{Minerals} \bibinfo{volume}{12},
  \bibinfo{pages}{526--544}.
\bibitem[{Tian et~al.(2019)Tian, Shen, Chen and He}]{FCOS}
\bibinfo{author}{Tian, Z.}, \bibinfo{author}{Shen, C.}, \bibinfo{author}{Chen,
  H.}, \bibinfo{author}{He, T.}, \bibinfo{year}{2019}.
\newblock \bibinfo{title}{Fcos: Fully convolutional one-stage object
  detection}, in: \bibinfo{booktitle}{IEEE/CVF International Conference on
  Computer Vision}, pp. \bibinfo{pages}{9626--9635}.
\bibitem[{Wang et~al.(2018)Wang, Zhang and Shao}]{particle}
\bibinfo{author}{Wang, R.}, \bibinfo{author}{Zhang, W.}, \bibinfo{author}{Shao,
  L.}, \bibinfo{year}{2018}.
\newblock \bibinfo{title}{Research of ore particle size detection based on
  image processing}, in: \bibinfo{booktitle}{Lecture Notes in Electrical
  Engineering}, pp. \bibinfo{pages}{505 -- 514}.
\bibitem[{Wang et~al.(2022)Wang, Xie, Li, Fan, Song, Liang, Lu, Luo and
  Shao}]{PVTv2}
\bibinfo{author}{Wang, W.}, \bibinfo{author}{Xie, E.}, \bibinfo{author}{Li,
  X.}, \bibinfo{author}{Fan, D.P.}, \bibinfo{author}{Song, K.},
  \bibinfo{author}{Liang, D.}, \bibinfo{author}{Lu, T.}, \bibinfo{author}{Luo,
  P.}, \bibinfo{author}{Shao, L.}, \bibinfo{year}{2022}.
\newblock \bibinfo{title}{Pvt v2: Improved baselines with pyramid vision
  transformer}.
\newblock \bibinfo{journal}{Computational Visual Media} \bibinfo{volume}{8},
  \bibinfo{pages}{415--424}.
\bibitem[{Wang et~al.(2020)Wang, Huang, Darrell, Gonzalez and Yu}]{TFA}
\bibinfo{author}{Wang, X.}, \bibinfo{author}{Huang, T.E.},
  \bibinfo{author}{Darrell, T.}, \bibinfo{author}{Gonzalez, J.E.},
  \bibinfo{author}{Yu, F.}, \bibinfo{year}{2020}.
\newblock \bibinfo{title}{Frustratingly simple few-shot object detection}, in:
  \bibinfo{booktitle}{Proceedings of the 37th International Conference on
  Machine Learning}, pp. \bibinfo{pages}{9919--9928}.
\bibitem[{Wang et~al.(2019)Wang, Ramanan and Hebert}]{MetaLearning}
\bibinfo{author}{Wang, Y.X.}, \bibinfo{author}{Ramanan, D.},
  \bibinfo{author}{Hebert, M.}, \bibinfo{year}{2019}.
\newblock \bibinfo{title}{Meta-learning to detect rare objects}, in:
  \bibinfo{booktitle}{IEEE/CVF International Conference on Computer Vision},
  pp. \bibinfo{pages}{9924--9933}.
\bibitem[{Woo et~al.(2018)Woo, Park, Lee and Kweon}]{CBAM}
\bibinfo{author}{Woo, S.}, \bibinfo{author}{Park, J.}, \bibinfo{author}{Lee,
  J.Y.}, \bibinfo{author}{Kweon, I.S.}, \bibinfo{year}{2018}.
\newblock \bibinfo{title}{Cbam: Convolutional block attention module}, in:
  \bibinfo{booktitle}{European Conference on Computer Vision}, pp.
  \bibinfo{pages}{3--19}.
\bibitem[{Wu et~al.(2020)Wu, Liu, Huang and Wang}]{MPSR}
\bibinfo{author}{Wu, J.}, \bibinfo{author}{Liu, S.}, \bibinfo{author}{Huang,
  D.}, \bibinfo{author}{Wang, Y.}, \bibinfo{year}{2020}.
\newblock \bibinfo{title}{Multi-scale positive sample refinement for few-shot
  object detection}, in: \bibinfo{booktitle}{European Conference on Computer
  Vision}, pp. \bibinfo{pages}{456--472}.
\bibitem[{Xiao and Marlet(2020)}]{FSDetView}
\bibinfo{author}{Xiao, Y.}, \bibinfo{author}{Marlet, R.}, \bibinfo{year}{2020}.
\newblock \bibinfo{title}{Few-shot object detection and viewpoint estimation
  for objects in the wild}, in: \bibinfo{booktitle}{European Conference on
  Computer Vision}, pp. \bibinfo{pages}{192--210}.
\bibitem[{Yan et~al.(2019)Yan, Chen, Xu, Wang, Liang and Lin}]{Meta_RCNN}
\bibinfo{author}{Yan, X.}, \bibinfo{author}{Chen, Z.}, \bibinfo{author}{Xu,
  A.}, \bibinfo{author}{Wang, X.}, \bibinfo{author}{Liang, X.},
  \bibinfo{author}{Lin, L.}, \bibinfo{year}{2019}.
\newblock \bibinfo{title}{Meta r-cnn: Towards general solver for instance-level
  low-shot learning}, in: \bibinfo{booktitle}{IEEE/CVF International Conference
  on Computer Vision}, pp. \bibinfo{pages}{9576--9585}.
\bibitem[{Yu et~al.(2018)Yu, Wang and Darrell}]{dla}
\bibinfo{author}{Yu, F.}, \bibinfo{author}{Wang, D.}, \bibinfo{author}{Darrell,
  T.}, \bibinfo{year}{2018}.
\newblock \bibinfo{title}{Deep layer aggregation}, pp.
  \bibinfo{pages}{2403--2412}.
\bibitem[{Zhang et~al.(2021)Zhang, Wang, Dayoub and Sünderhauf}]{VarifocalNet}
\bibinfo{author}{Zhang, H.}, \bibinfo{author}{Wang, Y.},
  \bibinfo{author}{Dayoub, F.}, \bibinfo{author}{Sünderhauf, N.},
  \bibinfo{year}{2021}.
\newblock \bibinfo{title}{Varifocalnet: An iou-aware dense object detector},
  in: \bibinfo{booktitle}{IEEE/CVF Conference on Computer Vision and Pattern
  Recognition}, pp. \bibinfo{pages}{8510--8519}.
\bibitem[{Zhou et~al.(2021)Zhou, Koltun and Kr{\"a}henb{\"u}hl}]{centernet2}
\bibinfo{author}{Zhou, X.}, \bibinfo{author}{Koltun, V.},
  \bibinfo{author}{Kr{\"a}henb{\"u}hl, P.}, \bibinfo{year}{2021}.
\newblock \bibinfo{title}{Probabilistic two-stage detection}.
\newblock \bibinfo{journal}{arXiv preprint arXiv:2103.07461} .
\bibitem[{Zhu et~al.(2022)Zhu, Zhu, Wang, Zhang and Zhao}]{Multigranularity}
\bibinfo{author}{Zhu, P.}, \bibinfo{author}{Zhu, Z.}, \bibinfo{author}{Wang,
  Y.}, \bibinfo{author}{Zhang, J.}, \bibinfo{author}{Zhao, S.},
  \bibinfo{year}{2022}.
\newblock \bibinfo{title}{Multi-granularity episodic contrastive learning for
  few-shot learning}.
\newblock \bibinfo{journal}{Pattern Recognition} \bibinfo{volume}{131},
  \bibinfo{pages}{108820}.

\end{thebibliography}
